\documentclass[journal]{IEEEtran}
\usepackage{graphicx}
\usepackage{amsmath, amssymb}
\usepackage{soul, xcolor}
\usepackage{caption, subcaption}
\usepackage{hyperref}
\usepackage{cite}
\usepackage{url}
\usepackage{footnote}
\makesavenoteenv{tabular}
\usepackage{multirow}
\usepackage{array}

\begin{document}
\title{Dynamic Memory Transformer for Hyperspectral Image Classification}
\author{Muhammad Ahmad
\thanks{M. Ahmad is with SDAIA-KFUPM, Joint Research Center for Artificial Intelligence (JRCAI), King Fahd University of Petroleum and Minerals, Dhahran, 31261, Saudi Arabia. (e-mail: mahmad00@gmail.com).}
}
\markboth{Journal of \LaTeX\ }%
{M.Ahmad \MakeLowercase{\textit{et al.}}}
\maketitle
\begin{abstract}
Hyperspectral image (HSI) classification (HSIC) requires effective modeling of complex spatial-spectral dependencies under limited labeled data and high dimensionality. While transformer-based models have shown strong capability in capturing long-range contextual information, they often introduce redundant attention patterns, which limits their effectiveness for fine-grained HSI analysis. To address these challenges, this paper proposes MemFormer, a lightweight transformer architecture for HSIC that incorporates a dynamic memory-enhanced attention mechanism. The proposed design augments multi-head self-attention with a compact global memory module that progressively aggregates contextual information across layers, enabling efficient modeling of long-range dependencies while reducing attention redundancy. In addition, a Spatial-Spectral Positional Embedding (SSPE) is used to jointly encode spatial continuity and spectral ordering, providing structurally consistent representations without relying on convolution-based positional encodings. Extensive experiments conducted on three benchmark hyperspectral datasets, including Indian Pines, WHU-Hi-HanChuan, and WHU-Hi-HongHu, demonstrate that MemFormer achieves superior classification performance compared to representative convolutional, hybrid, and transformer-based methods. On the Indian Pines dataset, MemFormer attains an overall accuracy of up to 99.55\%, average accuracy of 99.38\%, and a $\kappa$ coefficient of 99.49\%, highlighting its effectiveness and efficiency for HSIC.
\end{abstract}
\section{Introduction}
\label{sec:intro}

Hyperspectral imaging (HSI) captures scenes across hundreds of narrow and contiguous spectral bands, enabling detailed characterization of material properties that is not possible with conventional RGB or multispectral imagery \cite{10820058, 11392780}. This rich spectral resolution has made HSI classification (HSIC) a key enabling technology in applications such as precision agriculture, environmental monitoring, mineral exploration, and Earth observation \cite{10035973, 11119702}. However, the same high dimensionality that makes HSI informative also introduces significant challenges for robust and efficient learning-based classification \cite{Ahmad18072025, 11322762, 11105087}.

In particular, HSIC is fundamentally constrained by three interrelated factors \cite{11037730}. First, spectral variability induced by illumination changes, atmospheric conditions, and sensor noise leads to large intra-class variance, which complicates discriminative modeling across scenes and acquisition conditions \cite{11321257, 9767615}. Second, HSIs exhibit complex spatial structures where local texture, object boundaries, and long-range contextual relationships jointly influence class semantics, requiring models to capture both fine-scale locality and global dependencies \cite{11361224, 11222092}. Third, the Hughes phenomenon arises due to the curse of dimensionality, where the number of spectral bands far exceeds the number of labeled samples, resulting in overfitting and degraded generalization performance when model complexity is not carefully controlled \cite{11222092, 11226902}. Consequently, an effective HSIC model must strike a delicate balance between representational power, computational efficiency, and the ability to jointly model spatial–spectral correlations under limited supervision \cite{9307220, li2024casformer, 11090003}.

Recent advances in transformer-based architectures have demonstrated strong potential for HSIC by leveraging self-attention mechanisms to model long-range dependencies directly from spectral–spatial tokens. For instance, Huang \emph{et al.} \cite{10677405} proposed SS-VFMT, which augments pre-trained vision foundation models with dedicated spectral and spatial modules, while Zhao \emph{et al.} \cite{10472541} introduced GSC-ViT, combining GroupWise separable convolution with multi-head self-attention to reduce computational overhead. Ahmad \emph{et al.} \cite{10604879} further developed SSFormer by employing implicit conditional positional encodings to improve adaptability to variable input sizes and spectral resolutions. Despite their effectiveness, these transformer-centric methods often rely on large-scale pre-training \cite{Ahmad03042025}, incur substantial computational and memory costs due to quadratic attention complexity \cite{rs13122275}, and exhibit limited robustness when transferred across datasets with differing spectral characteristics or spatial resolutions \cite{10767233, AHMAD2025130428, 9645266}.

To mitigate these issues, hybrid CNN–transformer architectures have been proposed to exploit the complementary strengths of convolutional inductive biases and global attention mechanisms. Yu \emph{et al.} \cite{yu2024hypersinet} introduced Hypersinet, which employs cross-attention to fuse spatial and spectral representations, while Ahmad \emph{et al.} \cite{10399798} proposed WaveFormer, leveraging wavelet transforms for more effective spectral–spatial downsampling. Although such hybrid designs enhance feature expressiveness, they significantly increase architectural complexity, training time, and sensitivity to hyperparameter choices. Additional variants, including PyFormer \cite{10681622}, CMT \cite{10443948}, DiffFormer \cite{10955699}, and MASSFormer \cite{10506482}, adopt hierarchical processing, multi-scale fusion, or memory-inspired components to improve representation learning. However, these models often struggle to capture fine-grained spectral–spatial interactions consistently, particularly under limited training data, and their memory or hierarchy designs can exacerbate overfitting.

Recent efforts have also focused on preserving structural integrity in hyperspectral representations. Zhang \emph{et al.} \cite{ZHANG2025111470} proposed the Tensor Transformer, which maintains spatial–spectral coherence through tensor-based self-attention, while Wang \emph{et al.} \cite{10843260} introduced a CNN–transformer hybrid augmented with graph contrastive learning for Mars HSIC. Although promising, such approaches are either computationally demanding or tailored to highly specific application domains, limiting their general applicability to diverse Earth observation scenarios \cite{ahmad2024multihead}.

Despite substantial progress several critical challenges remain unresolved: (\emph{i}) effectively balancing local feature extraction and global context modeling without introducing excessive architectural complexity \cite{10681548}; (\emph{ii}) reducing the high computational and memory overhead inherent to transformer-based attention mechanisms \cite{10599231,10419118}; (\emph{iii}) robustly integrating spatial and spectral information in a manner that generalizes across domains and sensor configurations \cite{10711882}; and (\emph{iv}) mitigating overfitting risks associated with complex hierarchical or memory-based designs under limited labeled data \cite{10745620}. These limitations motivate the need for a more efficient yet expressive modeling paradigm for HSIC \cite{10685113, 9903062}.

To address these challenges, this work proposes MemFormer, a dynamic memory-enhanced transformer architecture tailored for HSIC. The key idea is to introduce a compact global memory mechanism that complements self-attention by retaining and refining contextual information across layers, thereby reducing redundancy while preserving discriminative capacity. The main contributions of this work are summarized as follows:

\begin{itemize}
    \item \textbf{Memory-enhanced multi-head attention with dynamic refinement:} We integrate a global memory module directly into the attention mechanism, which is dynamically updated across transformer layers to capture long-range dependencies efficiently while controlling computational and memory costs.
    
    \item \textbf{Spatial–Spectral Positional Embedding (SSPE):} We design a domain-aware positional encoding that explicitly preserves spatial continuity and spectral ordering, enabling structural integrity without relying on convolution-heavy encodings or costly pre-processing.
\end{itemize}

The remainder of this paper is organized as follows. Section~\ref{sec:pm} details the proposed methodology. Section~\ref{sec:abla} presents ablation studies to analyze the contribution of each component. Section~\ref{sec:comp} provides comprehensive comparisons with state-of-the-art methods on benchmark hyperspectral datasets. Finally, Section~\ref{sec:concl} concludes the paper and outlines directions for future research.

\begin{figure*}[!hbt]
    \centering
    \includegraphics[width=0.99\linewidth]{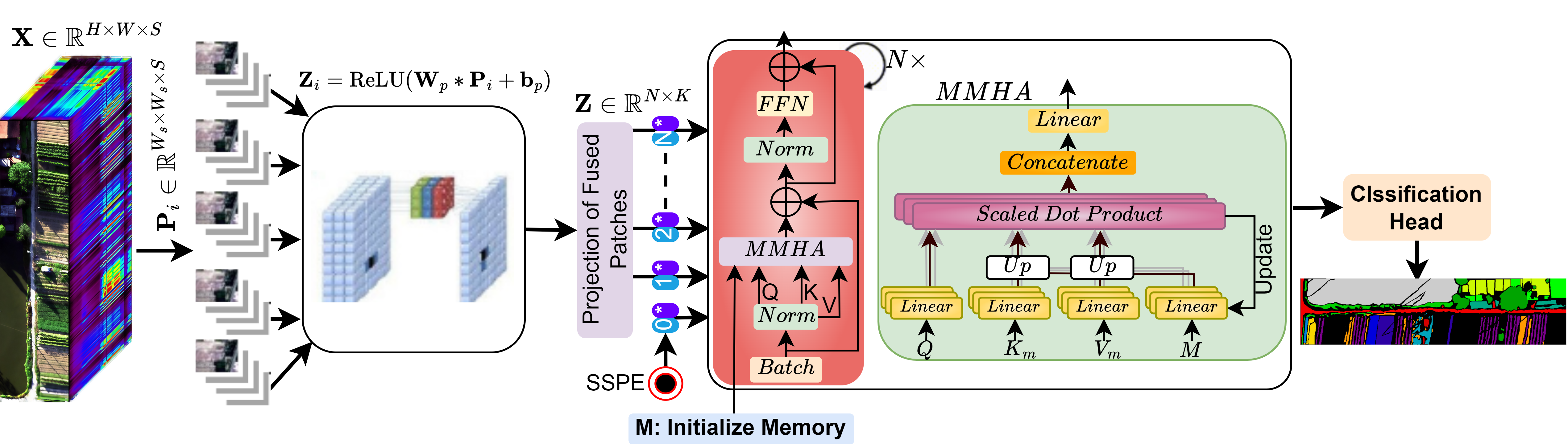}
    \caption{Overall architecture of dynamic memory-enhanced MHSA mechanism-based spatial-spectral transformer for HSIC.}
    \label{Fig1}
\end{figure*}

\section{Proposed Methodology}
\label{sec:pm}

Let the input HSI be denoted by $\mathbf{X} \in \mathbb{R}^{H \times W \times S}$, where $H$ and $W$ are the spatial dimensions and $S$ is the number of spectral bands. To reduce dimensionality while preserving local information, the HSI is split into $N$ non-overlapping \textit{patches} of size $W_s \times W_s \times S$, represented as $\mathbf{P}_i \in \mathbb{R}^{W_s \times W_s \times S}$, where $i = 1,2,\dots,N$ and $N = \tfrac{H \times W}{W_s^2}$. A shallow convolutional layer projects each patch into a feature embedding:

\begin{equation}
    \mathbf{Z}_i = \text{ReLU}(\mathbf{W}_p \ast \mathbf{P}_i + \mathbf{b}_p)
\end{equation}
where $\mathbf{W}_p \in \mathbb{R}^{K \times W_s \times W_s \times S}$ is the learnable kernel, $K$ is the embedding dimension, $\mathbf{b}_p$ is a bias term, and $\ast$ denotes convolution. After projection, embeddings from all patches are concatenated into $\mathbf{Z} \in \mathbb{R}^{N \times K}$.


Positional embeddings are crucial for distinguishing patches with similar spectra but different spatial locations. To this end, we introduce an SSPE that jointly encodes spatial coordinates and spectral order. Let $(x_i, y_i)$ denote the 2D coordinates of the $i$-th patch, and let $s_j$ denote the index of the $j$-th spectral band. The spatial encoding exploits sinusoidal functions:

\begin{align*}
E_x(x_i, 2j) &= \sin\!\left(\tfrac{x_i}{\lambda^{2j/d}}\right), \quad
E_x(x_i, 2j+1) = \cos\!\left(\tfrac{x_i}{\lambda^{2j/d}}\right) \\
E_y(y_i, 2j) &= \sin\!\left(\tfrac{y_i}{\lambda^{2j/d}}\right), \quad
E_y(y_i, 2j+1) = \cos\!\left(\tfrac{y_i}{\lambda^{2j/d}}\right)
\end{align*}
where $\lambda$ is a wavelength parameter and $d$ is the embedding size. The spatial encoding is then: 

\begin{equation}
    E_{\text{spatial}}(x_i, y_i) = [E_x(x_i); E_y(y_i)] \in \mathbb{R}^{K_s}
\end{equation}

The spectral encoding uses a similar sinusoidal scheme:

\begin{align}
E_{\text{spectral}}(s_j, 2k) = \sin\!\left(\tfrac{s_j}{\gamma^{2k/d}}\right) \\
E_{\text{spectral}}(s_j, 2k+1) = \cos\!\left(\tfrac{s_j}{\gamma^{2k/d}}\right)
\end{align}
where $\gamma$ is a scaling factor. The resulting embedding is $E_{\text{spectral}}(s_j) \in \mathbb{R}^{K_\sigma}$. 
For the sake of compatibility, both embeddings are projected to dimension $K$ via linear layers and concatenated:

\[
E_{\text{SSPE}} = \text{MLP}\Big(\big[\text{Proj}_s(E_{\text{spatial}}); \text{Proj}_\sigma(E_{\text{spectral}})\big]\Big) \in \mathbb{R}^{N \times K}
\]

This SSPE scheme allows the transformer to capture both spatial identity and continuity and spectral ordering. 

\subsection{Dynamic Memory-Enhanced Attention} 

Transformers typically rely on multi-head self-attention (MHSA), but standard MHSA can suffer from redundancy and information dilution. We enhance MHSA with a lightweight \textbf{dynamic memory mechanism} to retain contextual dependencies across layers. A classification token $\mathbf{z}_{\text{CLS}} \in \mathbb{R}^K$ is prepended to the embedding sequence, and SSPE is added:

\begin{equation}
    \mathbf{Z}' = [\mathbf{z}_{\text{CLS}}; \mathbf{Z}] + E_{\text{SSPE}}    
\end{equation}

Let the query matrix be $\mathbf{Q} \in \mathbb{R}^{B \times N \times K}$ for a batch size $B$. We maintain a global dynamic memory $\mathbf{M} \in \mathbb{R}^{M \times K}$, where $M$ is the memory length. This memory is non-trainable and updated dynamically per batch via a \textbf{First-In-First-Out (FIFO)} policy, which discards the oldest entries while preserving the most recent global features. 

To align it with the attention mechanism, the memory entries are projected into key and value spaces across the batches: 

\begin{align}
    \mathbf{K}_m &= \text{Tile}(\mathbf{M} \mathbf{W}_K, B) \\
    \mathbf{V}_m &= \text{Tile}(\mathbf{M} \mathbf{W}_V, B)
\end{align}
where $\mathbf{W}_K, \mathbf{W}_V \in \mathbb{R}^{K \times K}$ are learnable matrices, and $\text{Tile}(\cdot)$ replicates the memory across the batch dimension. This formulation results in query vectors derived from the current input tokens, while keys and values are sourced exclusively from the dynamic memory, enabling memory-conditioned attention.

Memory-based attention is computed as:

\begin{equation}
    \mathbf{A} = \text{Softmax}\!\left(\tfrac{\mathbf{Q}\mathbf{K}_m^\top}{\sqrt{K}}\right)\mathbf{V}_m
\end{equation}

The result is normalized via a residual connection:

\begin{equation}
    \mathbf{A}' = \text{LayerNorm}(\mathbf{Q} + \mathbf{A})
\end{equation}

After each batch, memory is updated with the averaged attention responses:

\begin{equation}
    \mathbf{m}_{\text{new}} = \tfrac{1}{B}\sum_{b=1}^{B}\Big(\tfrac{1}{N}\sum_{n=1}^{N} \mathbf{A}[b,n,:]\Big)
\end{equation}

This vector is appended, and the oldest entry is discarded:

\begin{equation}
    \mathbf{M} \leftarrow [\mathbf{M}[1:, :]; \mathbf{m}_{\text{new}}]
\end{equation}

Given the input embedding sequence $\mathbf{Z}' \in \mathbb{R}^{B \times (N+1) \times K}$, the query vectors are obtained via a linear projection $\mathbf{Q} = \mathbf{Z}' \mathbf{W}_Q$, where $\mathbf{W}_Q \in \mathbb{R}^{K \times K}$ is a learnable projection matrix. This operation enables each token to formulate a query representing its information requirement with respect to the global context. With the updated memory, attention is recomputed:

\begin{equation}
    \mathbf{A}_{\text{final}} = \text{Softmax}\!\left(\tfrac{\mathbf{Q}\mathbf{K}_m^{\text{updated}\top}}{\sqrt{K}}\right)\mathbf{V}_m^{\text{updated}}
\end{equation}

The final normalized embedding is:

\begin{equation}
    \mathbf{A}_{\text{final}}' = \text{LayerNorm}(\mathbf{Q} + \mathbf{A}_{\text{final}})
\end{equation}

\textbf{Feedforward and Classification:} A position-wise feedforward network introduces non-linearity:

\begin{equation}
    \text{FFN}(\mathbf{X}) = \text{ReLU}(\mathbf{X}\mathbf{W}_1 + \mathbf{b}_1)\mathbf{W}_2 + \mathbf{b}_2
\end{equation}
where $\mathbf{W}_1 \in \mathbb{R}^{K \times D}$, $\mathbf{W}_2 \in \mathbb{R}^{D \times K}$, and $D$ is the hidden dimension. The final representation is: 

\begin{equation}
    \mathbf{Z}'' = \text{LayerNorm}\big(\mathbf{A}_{\text{final}}' + \text{FFN}(\mathbf{A}_{\text{final}}')\big)
\end{equation}

The classification is performed via the CLS token:

\begin{equation}
    \mathbf{z}_{\text{CLS}}^{\text{final}} = \mathbf{Z}''[0, :], \quad \mathbf{y} = \text{Softmax}(\mathbf{z}_{\text{CLS}}^{\text{final}}\mathbf{W}_c + \mathbf{b}_c)
\end{equation}
where $\mathbf{W}_c \in \mathbb{R}^{K \times C}$ and $C$ is the number of classes.

\section{Ablation Study}
\label{sec:abla}

To validate the effectiveness of individual components in our proposed MemFormer, we conduct an ablation study on three widely used HSI datasets: Indian Pines (IP), WHU-Hi-HanChuan (HC), and WHU-Hi-HongHu (HH). We specifically examine three questions: Does the proposed memory-enhanced attention improve classification compared to standard self-attention? How does our SSPE compare with other positional encoding strategies?  What is the optimal memory size for balancing context preservation and redundancy?

\begin{table}[!hbt]
    \centering
    \caption{Impact of memory-enhanced attention compared to standard self-attention across three datasets.}
    \resizebox{\columnwidth}{!}{\begin{tabular}{c|ccc|ccc|ccc} \hline 
        \multirow{2}{*}{\textbf{Attention}} & \multicolumn{3}{c|}{\textbf{IP}} & \multicolumn{3}{c|}{\textbf{HC}} & \multicolumn{3}{c}{\textbf{HH}} \\ \cline{2-10}
        & $\kappa$ & OA & AA & $\kappa$ & OA & AA & $\kappa$ & OA & AA \\ \hline 
        Standard & 94.93 & 95.55 & 95.51 & 97.29 & 97.68 & 95.59 & 96.22 & 97.01 & 92.85 \\
        Memory   & \textbf{99.49} & \textbf{99.55} & \textbf{99.38} & \textbf{99.14} & \textbf{99.27} & \textbf{98.67} & \textbf{98.12} & \textbf{98.51} & \textbf{97.26} \\ \hline 
    \end{tabular}}
    \label{Tab1}
\end{table}

\textbf{Effect of Memory Attention:} We first compare the proposed memory-augmented attention with the standard self-attention mechanism. Table \ref{Tab1} reports the results. \textbf{Observation:} Incorporating memory substantially improves performance across all datasets. On IP, the $\kappa$ score increases by more than 4\%, while on HC and HH, improvements of nearly 2\% are observed. These results confirm that the memory module helps retain long-range contextual information critical for hyperspectral data.

\begin{table}[!hbt]
   \centering
    \caption{Comparison of different positional encoding mechanisms. The 
    SSPE consistently yields the highest accuracy.}
    \resizebox{\columnwidth}{!}{\begin{tabular}{c|ccc|ccc|ccc} \hline 
        \multirow{2}{*}{\textbf{PE}} & \multicolumn{3}{c|}{\textbf{IP}} & \multicolumn{3}{c|}{\textbf{HC}} & \multicolumn{3}{c}{\textbf{HH}} \\ \cline{2-10}
        & $\kappa$ & OA & AA & $\kappa$ & OA & AA & $\kappa$ & OA & AA \\ \hline 
        None        & 98.22 & 98.41 & 97.85 & 98.00 & 98.36 & 96.50 & 97.55 & 97.90 & 95.12 \\
        Learnable   & 98.74 & 98.90 & 98.30 & 98.54 & 98.76 & 97.36 & 97.99 & 98.33 & 95.88 \\
        Sinusoidal  & 98.53 & 98.72 & 98.09 & 98.36 & 98.64 & 97.11 & 97.81 & 98.19 & 95.63 \\
        SSPE        & \textbf{99.01} & \textbf{99.10} & \textbf{99.72} & \textbf{98.87} & \textbf{99.01} & \textbf{98.92} & \textbf{98.27} & \textbf{98.60} & \textbf{97.77} \\ \hline 
    \end{tabular}}
    \label{PosEnc}
\end{table}

\textbf{Effect of Positional Encoding Strategies:} Next, we assess the importance of positional embeddings. We compare four strategies: (i) no positional encoding, (ii) learnable embedding, (iii) sinusoidal embedding, and (iv) the proposed SSPE. Results are presented in Table \ref{PosEnc}. \textbf{Observation:} The absence of positional encoding causes a clear performance drop, particularly in average accuracy (AA). While learnable and sinusoidal encodings partially mitigate this issue, SSPE consistently outperforms all alternatives. This demonstrates the value of domain-aware positional encoding tailored to spectral–spatial properties.

\begin{table}[!hbt]
    \centering
    \caption{Effect of memory size on accuracy (OA).}
    \resizebox{\columnwidth}{!}{\begin{tabular}{c|c|c|c|c|c|c|c}
        \hline
        \textbf{Dataset} & \textbf{1} & \textbf{5} & \textbf{10} & \textbf{15} & \textbf{20} & \textbf{25} & \textbf{30} \\
        \hline
        IP & 98.16 & 95.80 & \textbf{99.23} & 91.18 & 98.65 & 98.88 & 98.67 \\
        HH & 97.45 & 97.25 & \textbf{98.51} & 97.64 & 97.78 & 97.22 & 98.02 \\
        HC & 97.56 & 97.28 & \textbf{99.26} & 96.45 & 97.54 & 97.59 & 97.44 \\
        \hline
    \end{tabular}}
    \label{Memory}
\end{table}

\textbf{Effect of Memory Size:} Finally, we study how memory size (number of tokens stored) impacts performance. Table \ref{Memory} shows the overall accuracy (OA) when varying the memory length from 1 to 30. \textbf{Observation:} A memory length of around 10 consistently delivers the best or near-best performance across all datasets. Very small memory (e.g., 1 token) fails to retain sufficient context, while excessively large memory (e.g., 25–30 tokens) introduces redundancy and noise, leading to performance degradation. This analysis highlights the importance of balancing memory size to preserve informative context without overwhelming the attention mechanism.

\section{Comparative Results and Discussion}
\label{sec:comp}

We evaluate MemFormer (MSST) against several representative baselines: 2D CNN \cite{hu2015deep}, 3D CNN \cite{hamida2018deep}, Hybrid CNN (HCNN) \cite{paoletti2018new}, Pyramid Transformer (PF) \cite{10681622}, WaveFormer (WF) \cite{10399798}, Hyperspectral Spatial–Spectral Transformer (HSST) \cite{yu2024hypersinet}, and Spectral–Spatial Transformer (SST) \cite{10884842}. To ensure fairness, all models are trained with identical configurations: patch size $14$, embedding dimension $K=15$, 25\% data for training/validation, and 50\% for testing (stratified cross-validation). Training uses the Adam optimizer for 50 epochs with learning rate $10^{-3}$ and weight decay $10^{-6}$. All transformer models use 4 layers, 8 heads, and a dropout rate $0.1$.

\begin{figure*}
    \centering
    \includegraphics[width=0.95\linewidth]{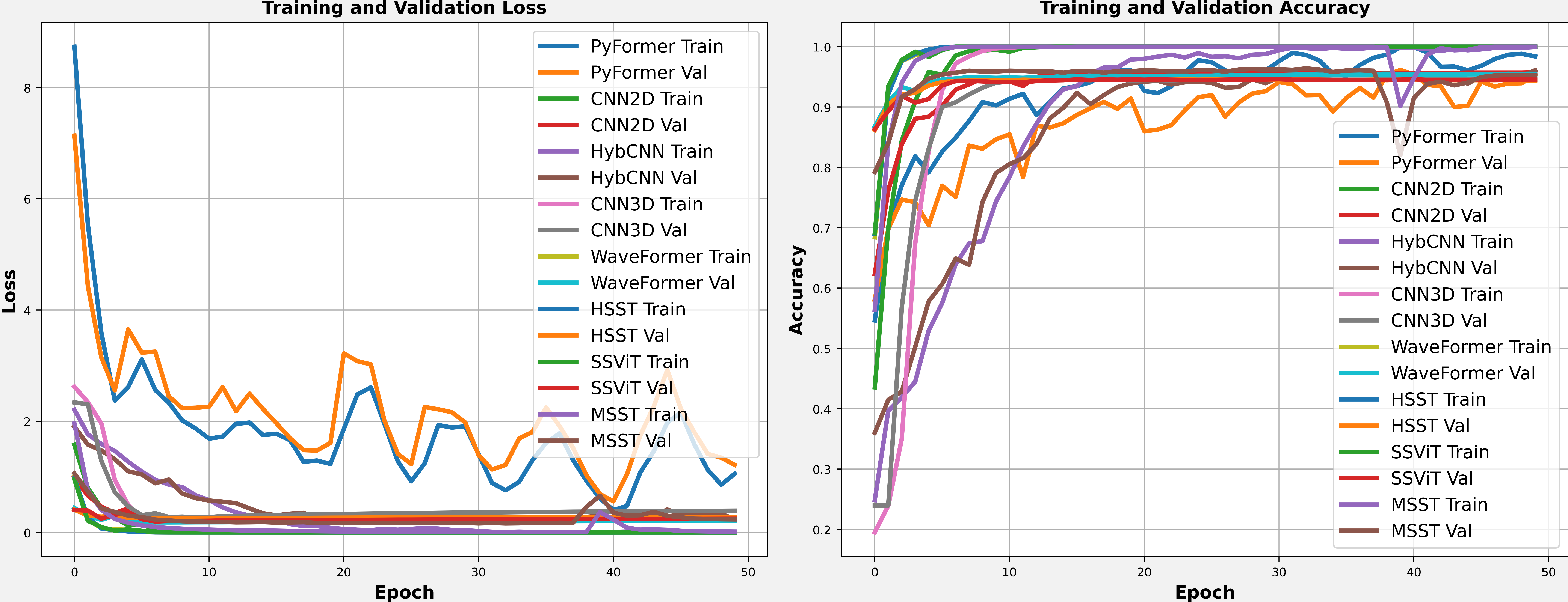}
    \caption{Accuracy and loss trends for training and validation samples on the IP dataset across 50 epochs.}
    \label{Loss}
\end{figure*}

\begin{table}[!hbt]
    \centering
    \caption{\textbf{IP dataset:} Per-class classification results with overall accuracy (OA), average accuracy (AA), and $\kappa$. MSST achieves the best performance across most classes.}
    \resizebox{\columnwidth}{!}{\begin{tabular}{c|cccccccc} \hline 
        \textbf{Class} & \textbf{2DCNN} & \textbf{HCNN} & \textbf{3DCNN} & \textbf{HSST} & \textbf{SST} & \textbf{PF} & \textbf{WF} & \textbf{MSST} \\ \hline 
            1 & 56.5217 & 60.8695 & 73.9130 & 69.5652 & \textbf{100.0000} & 91.3043 & 91.3043 & 95.6521 \\
            2 & 93.8375 & 97.3389 & 93.9775 & 92.0168 & 92.4369 & 95.2380 & 92.2969 & \textbf{99.2997} \\
            3 & 96.6265 & 96.3855 & 96.6265 & 95.6626 & 96.3855 & 91.5662 & 96.8674 & \textbf{100.0000} \\
            4 & 85.5932 & 90.6779 & 83.0508 & 94.0677 & 92.3728 & 89.8305 & 86.4406 & \textbf{100.0000} \\
            5 & 96.2809 & 97.5206 & 95.0413 & \textbf{99.1735} & 97.5206 & 96.2809 & 97.1074 & 97.5206 \\
            6 & 98.6301 & \textbf{99.7260} & 99.4520 & 99.1780 & 98.9041 & 98.6301 & 98.9041 & 99.1780 \\
            7 & \textbf{100.0000} & 71.4285 & 92.8571 & \textbf{100.0000} & \textbf{100.0000} & 92.8571 & \textbf{100.0000} & \textbf{100.0000} \\
            8 & \textbf{100.0000} & 99.5815 & 99.5815 & \textbf{100.0000} & \textbf{100.0000} & \textbf{100.0000} & \textbf{100.0000} & \textbf{100.0000} \\
            9 & 60.0000 & 50.0000 & \textbf{100.0000} & 90.0000 & 80.0000 & 90.0000 & 80.0000 & \textbf{100.0000} \\
            10 & 90.9465 & 95.6790 & 91.9753 & 86.8312 & 87.6543 & 89.5061 & 85.1851 & \textbf{99.3827} \\
            11 & 96.8241 & 98.8599 & 97.3127 & 96.3355 & 96.8241 & 98.0456 & 96.9055 & \textbf{99.7557} \\
            12 & 96.2962 & 96.6329 & 91.2457 & 90.9090 & 94.2760 & 92.2558 & 88.2154 & \textbf{99.3265} \\
            13 & 97.0588 & 97.0588 & 99.0196 & \textbf{100.0000} & 99.0196 & \textbf{100.0000} & \textbf{100.0000} & \textbf{100.0000} \\
            14 & 99.2101 & 99.0521 & 98.8941 & 98.5781 & 98.4202 & 99.0521 & 97.7883 & \textbf{100.0000} \\
            15 & 95.8549 & 83.4196 & 93.2642 & 97.4093 & 94.3005 & 97.4093 & 97.9274 & \textbf{100.0000} \\
            16 & \textbf{100.0000} & 93.4782 & \textbf{100.0000} & \textbf{100.0000} & \textbf{100.0000} & 93.4782 & \textbf{100.0000} & \textbf{100.0000} \\ \hline
        \textbf{Para} & 1078336 & 1380480 & 15808224 & 836816 & 836816 & 16680662 & 2667408 & 847632 \\ \hline

        \textbf{Train (s)} & 24.03 & 30.83 & 40.96 & 102.46 & 106.02 & 1019.90 & 96.38 & 101.64 \\ \hline
        \textbf{Test (s)} & 0.90 & 0.83 & 0.78 & 2.37 & 2.33 & 12.92 & 2.48 & 2.34 \\ \hline
        \textbf{Flops} & 2670080 & 18778112 & 15732736 & 110592 & 110592 & 15730688 & 167936 & 3108864 \\ \hline

        \textbf{$\kappa$} & 95.2557 & 96.4373 & 95.1454 & 94.5476 & 94.9254 & 95.2985 & 94.1671 & \textbf{99.4882} \\ \hline
        \textbf{OA} & 95.8439 & 96.8780 & 95.7463 & 95.2195 & 95.5512 & 95.8829 & 94.8878 & \textbf{99.5512} \\ \hline
        \textbf{AA} & 91.4800 & 89.2318 & 94.1382 & 94.3579 & 95.5071 & 94.7159 & 94.3089 & \textbf{99.3822} \\ \hline
    \end{tabular}}
    \label{IPT}
\end{table}
\begin{figure}[!hbt]
    \centering
    \begin{subfigure}{0.11\textwidth}
	\includegraphics[width=0.99\textwidth]{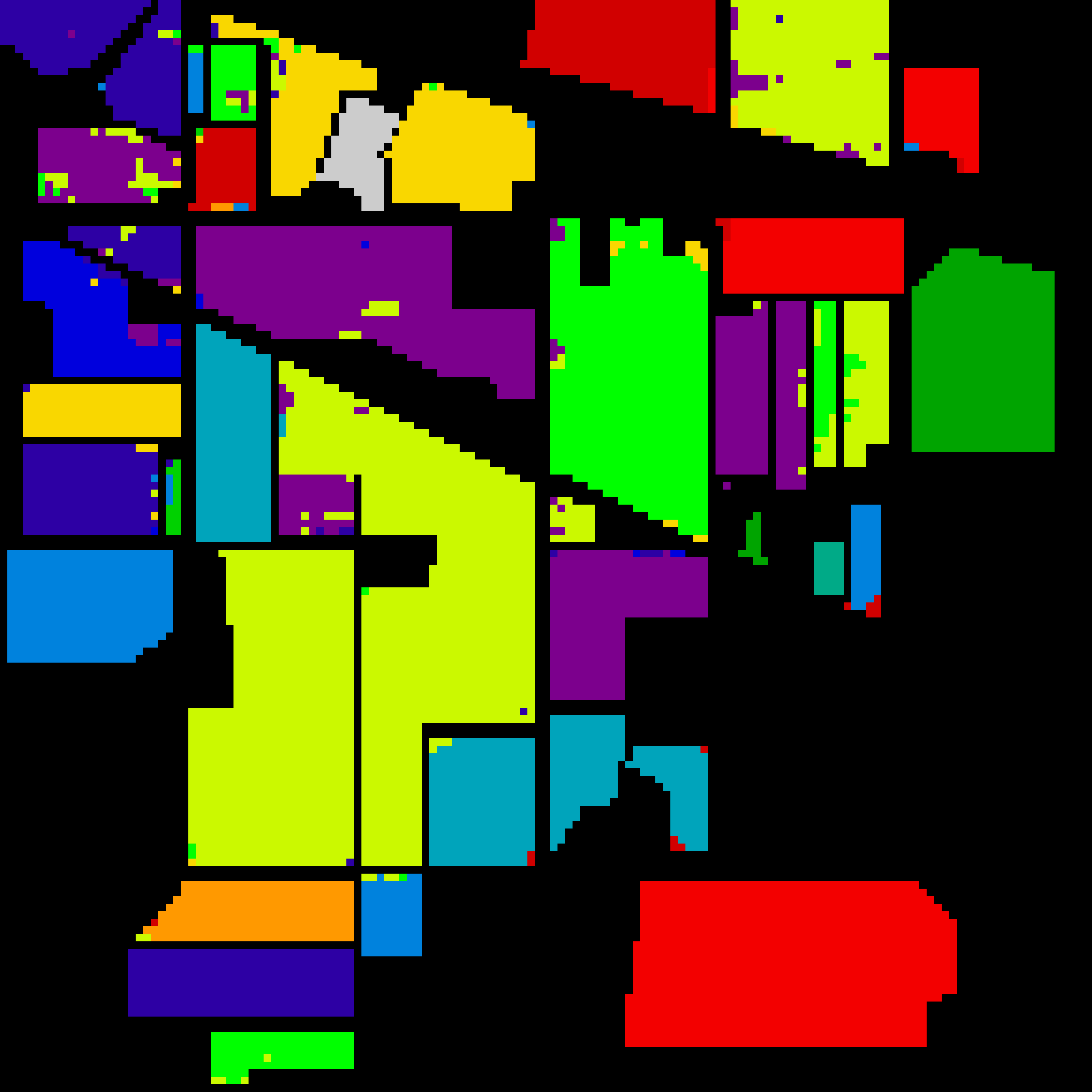}
	\caption*{2D} 
    \end{subfigure}
    \begin{subfigure}{0.11\textwidth}
	\includegraphics[width=0.99\textwidth]{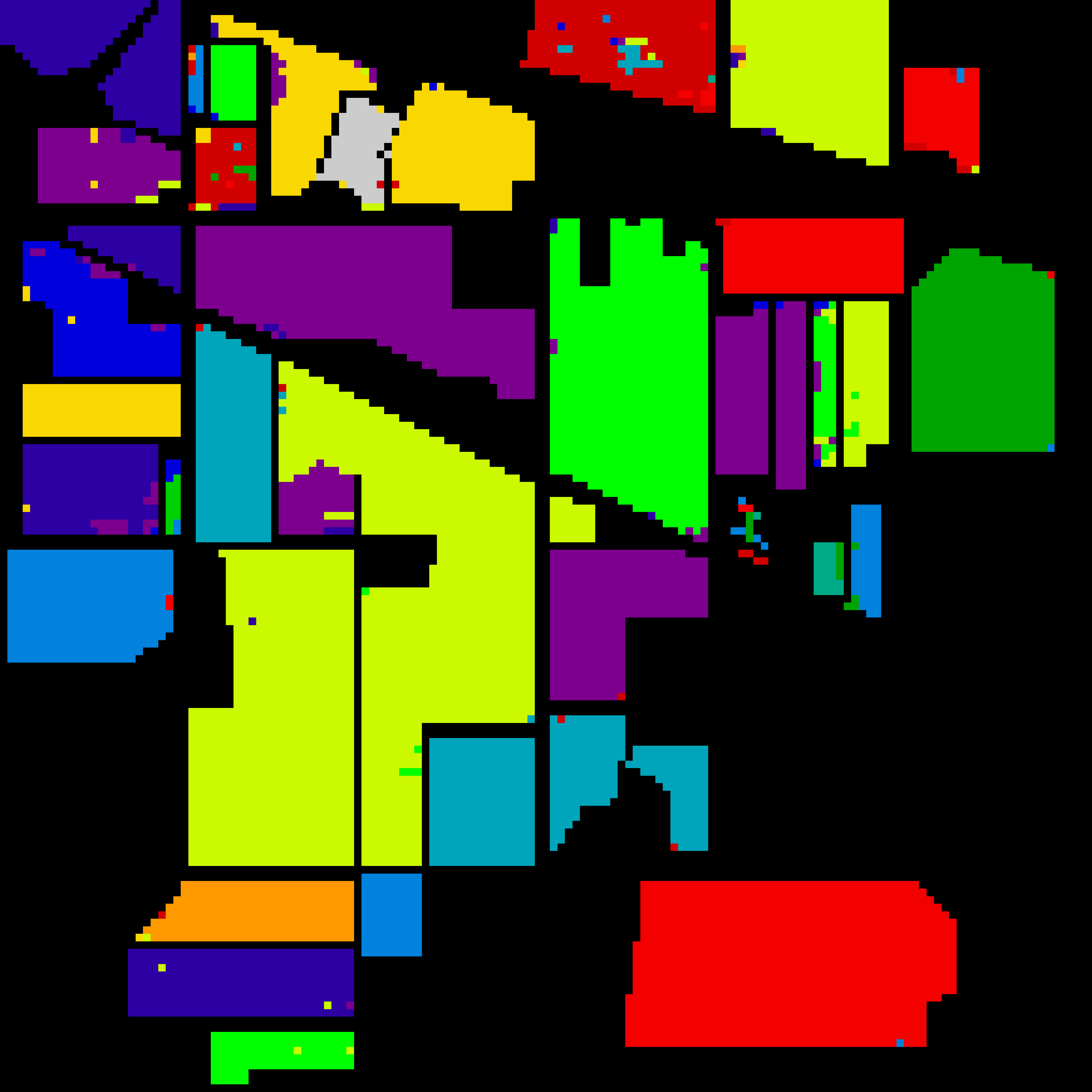}
	\caption*{Hyb}
    \end{subfigure}
    \begin{subfigure}{0.11\textwidth}
	\includegraphics[width=0.99\textwidth]{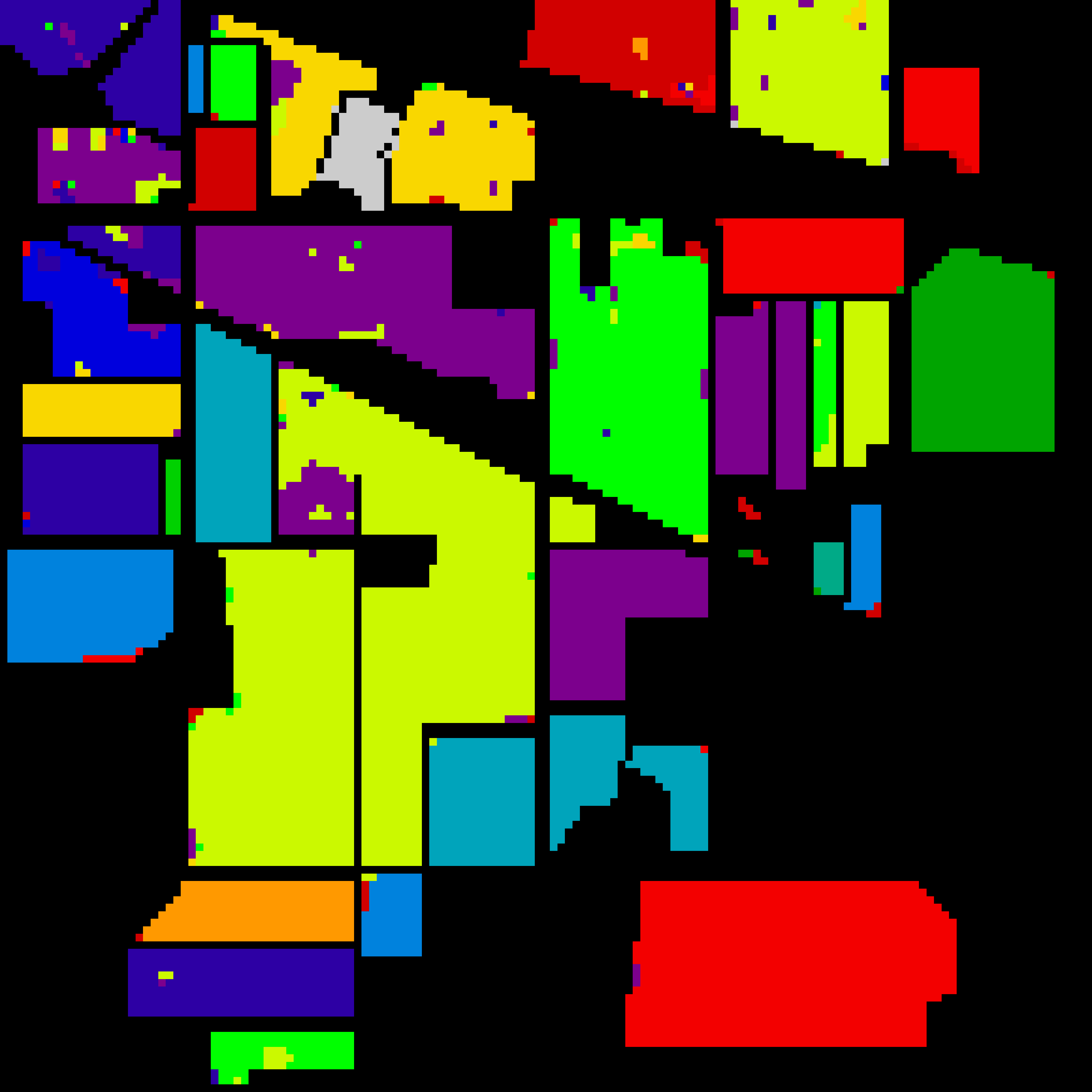}
	\caption*{3D}
    \end{subfigure} 
    \begin{subfigure}{0.11\textwidth}
	\includegraphics[width=0.99\textwidth]{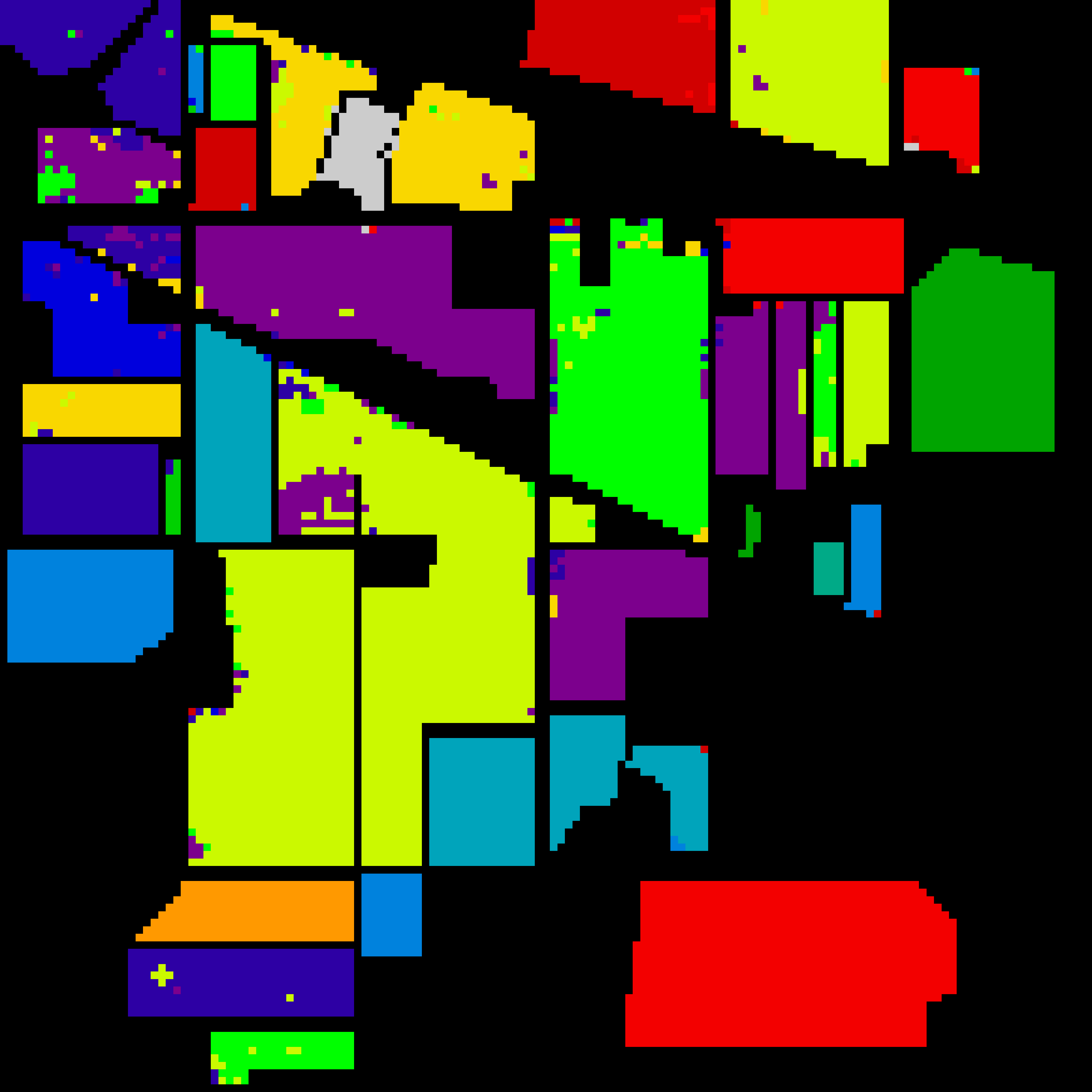}
	\caption*{HSST}
    \end{subfigure}
    \begin{subfigure}{0.11\textwidth}
	\includegraphics[width=0.99\textwidth]{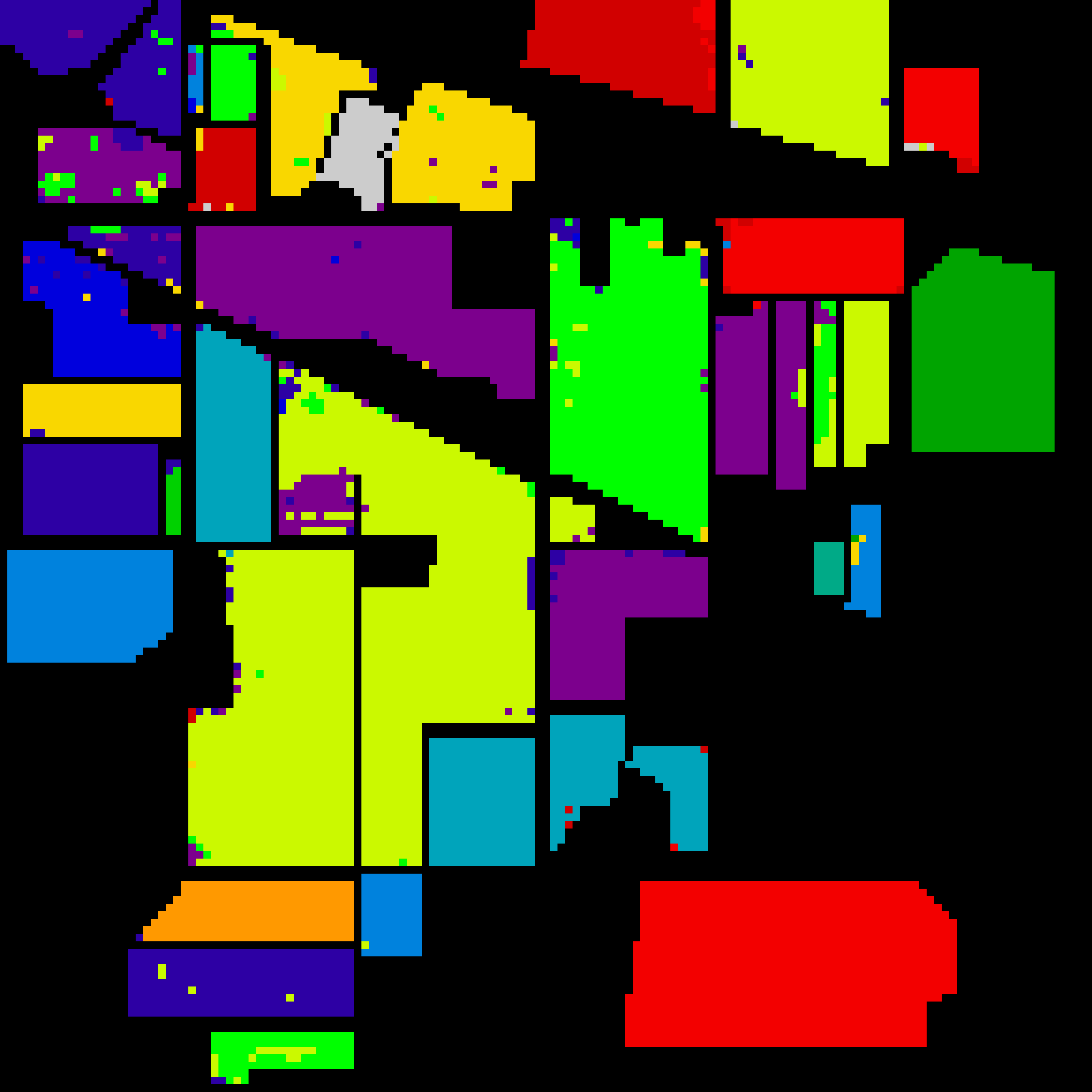}
	\caption*{SST}
    \end{subfigure}
    \begin{subfigure}{0.11\textwidth}
	\includegraphics[width=0.99\textwidth]{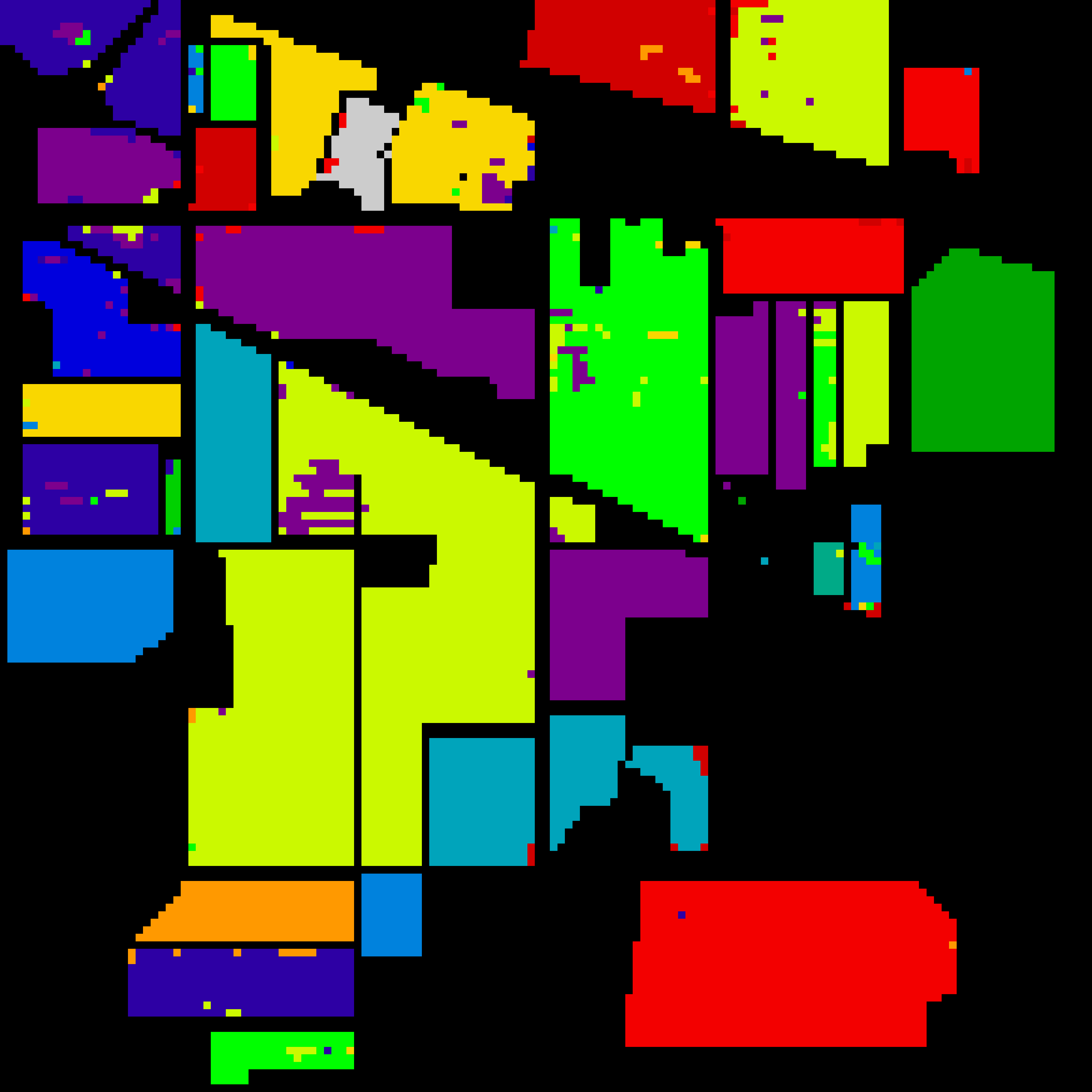}
	\caption*{PF}
    \end{subfigure}
    \begin{subfigure}{0.11\textwidth}
	\includegraphics[width=0.99\textwidth]{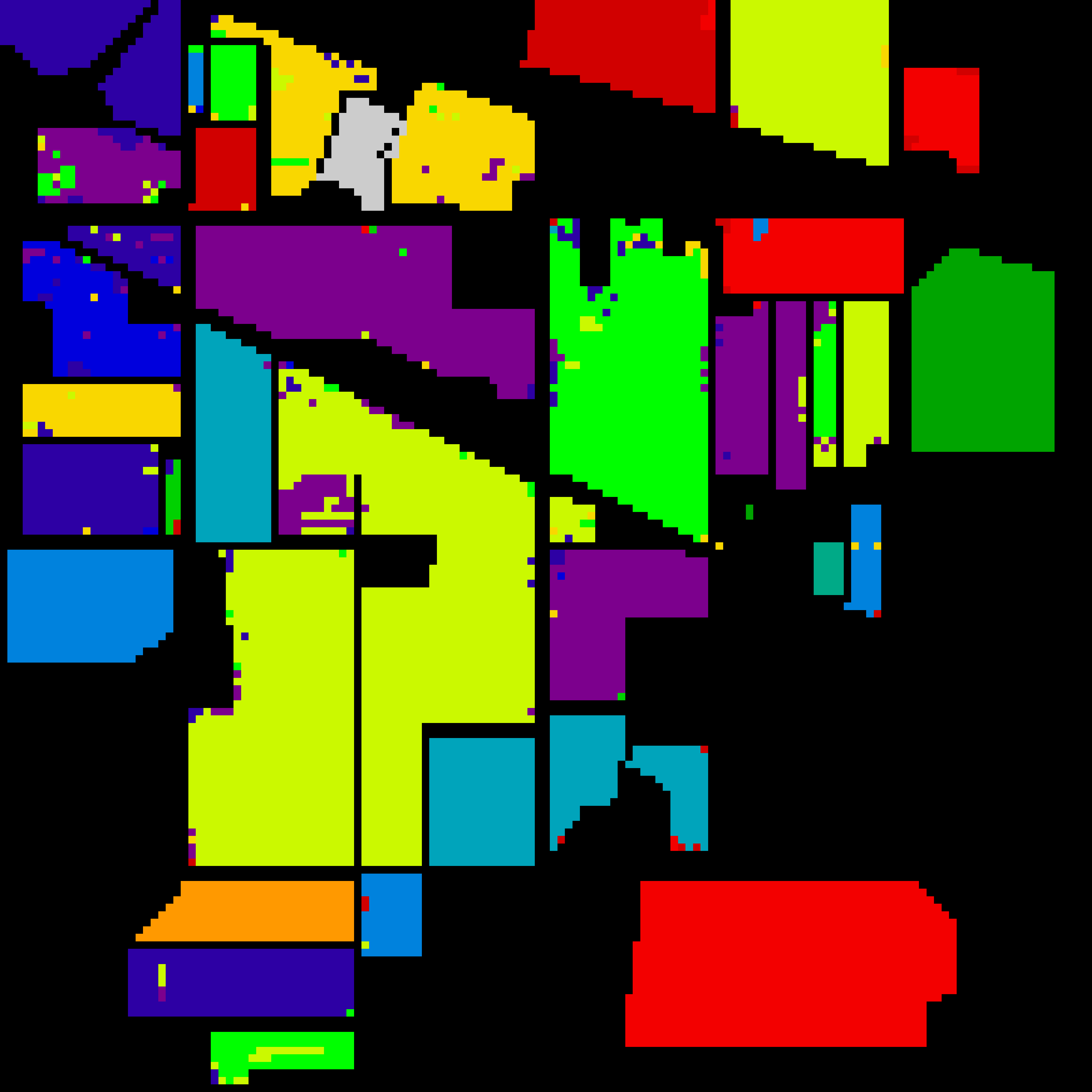}
	\caption*{WF}
    \end{subfigure}
    \begin{subfigure}{0.11\textwidth}
	\includegraphics[width=0.99\textwidth]{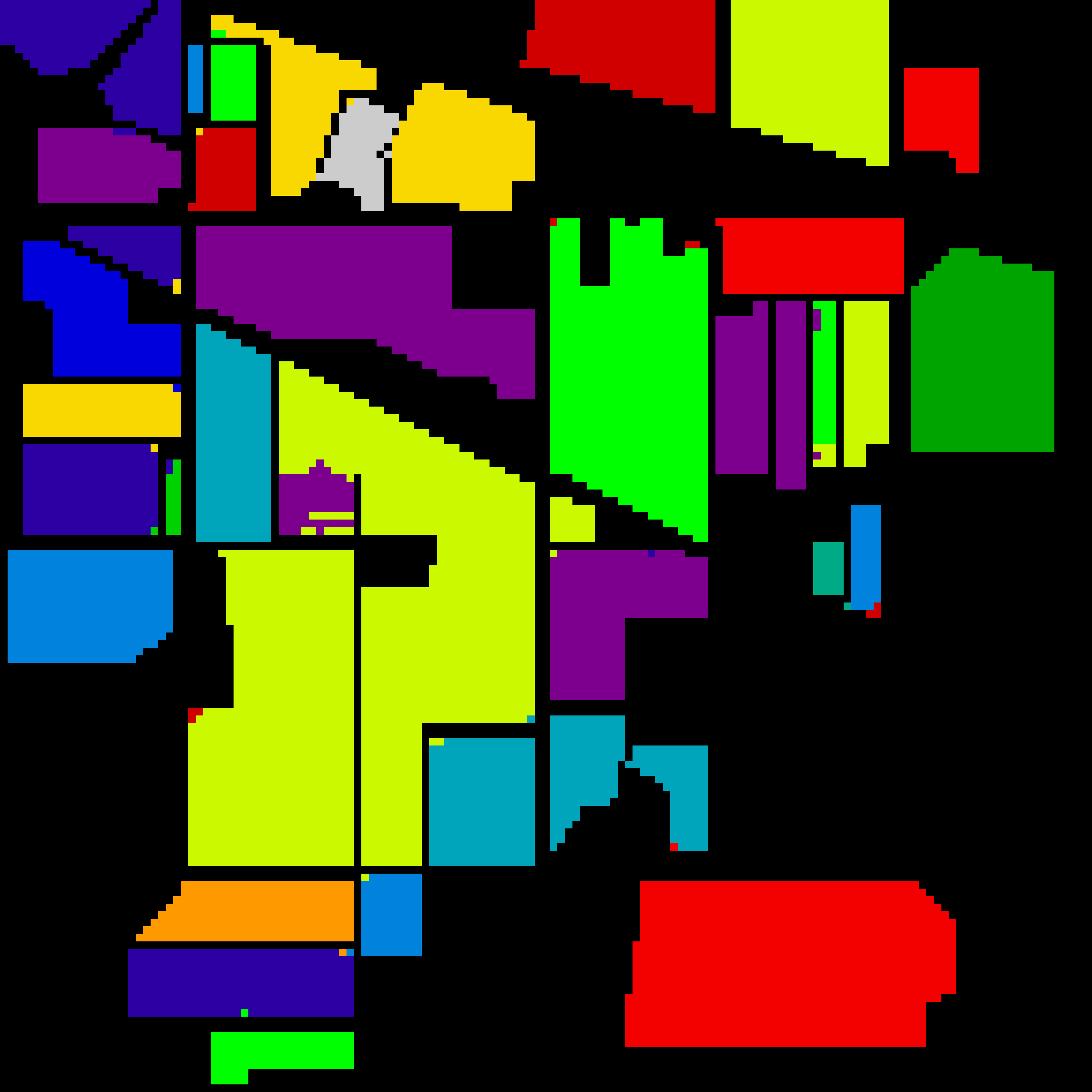}
	\caption*{MSST}
    \end{subfigure}
    \caption{\textbf{IP dataset:} Classification maps for different models, highlighting the superior spatial consistency of MSST.}
    \label{IPF}
    \vspace{-.2cm}
\end{figure}

The accuracy and loss trends for both training and validation are presented in Figure \ref{Loss}. To ensure a fair and rigorous evaluation, both qualitative and quantitative comparisons are conducted. The evaluation metrics include per-class classification performance, OA, AA, Kappa coefficient ($\kappa$), computational efficiency (FLOPs and runtime), and parameter count. Through this comparative analysis, we aim to demonstrate the effectiveness and superiority of the proposed MSST model in addressing the challenges of HSI classification.

\textbf{IP:} Table \ref{IPT} summarizes per-class results on IP, and Fig. \ref{IPF} shows the corresponding classification maps. \textbf{Observation:} MSST outperforms all baselines in terms of OA, AA, and $\kappa$. Importantly, classes that are difficult for CNNs and vanilla transformers (e.g., Class 2 and Class 10) are classified more accurately by MSST. The classification maps also reveal sharper boundaries and fewer misclassified patches.

\begin{figure}[!hbt]
        \centering
    \begin{subfigure}{0.11\textwidth}
	\includegraphics[width=0.99\textwidth]{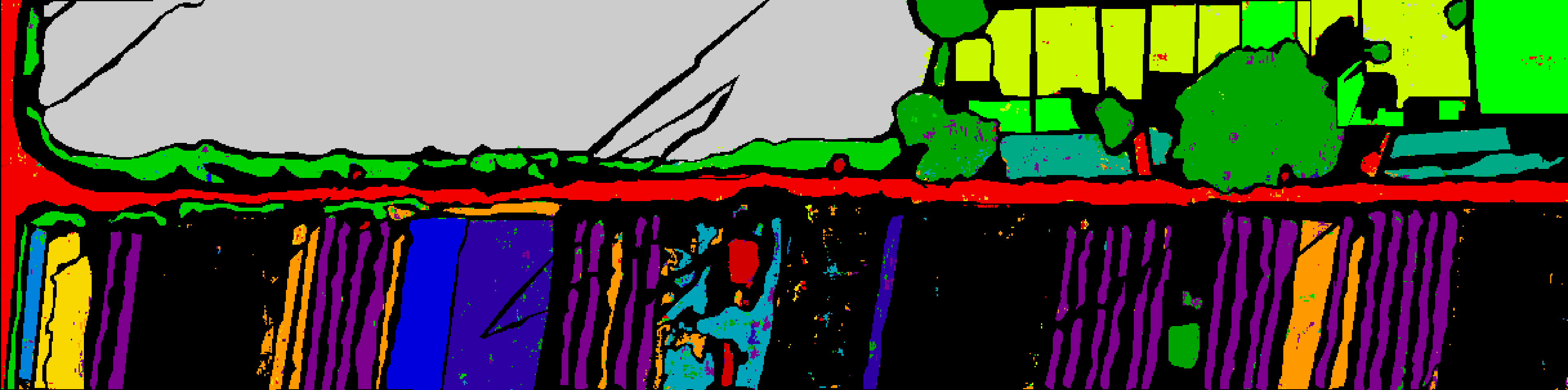}
	\caption*{2D} 
    \end{subfigure}
    \begin{subfigure}{0.11\textwidth}
	\includegraphics[width=0.99\textwidth]{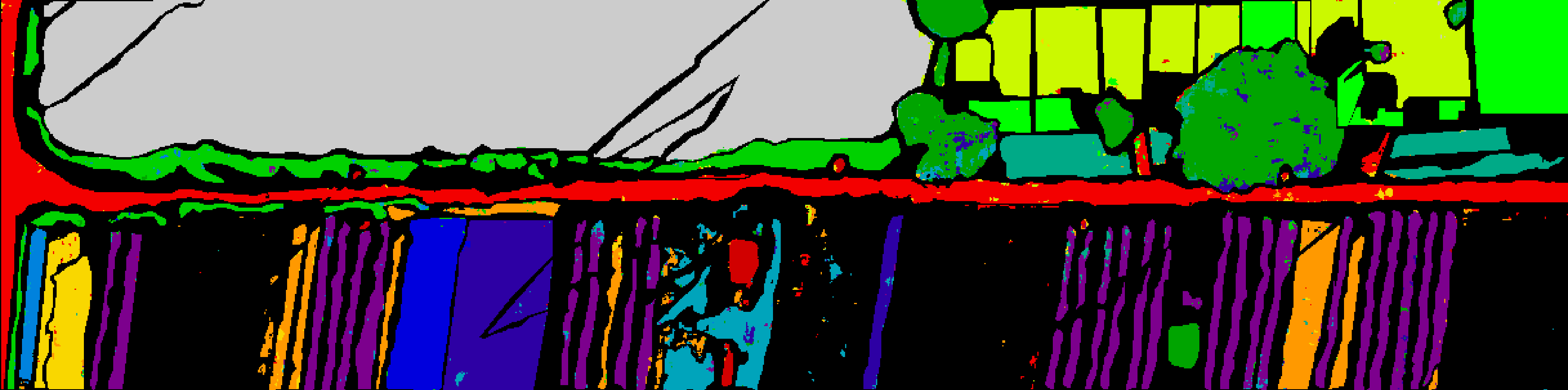}
	\caption*{Hyb}
    \end{subfigure}
    \begin{subfigure}{0.11\textwidth}
	\includegraphics[width=0.99\textwidth]{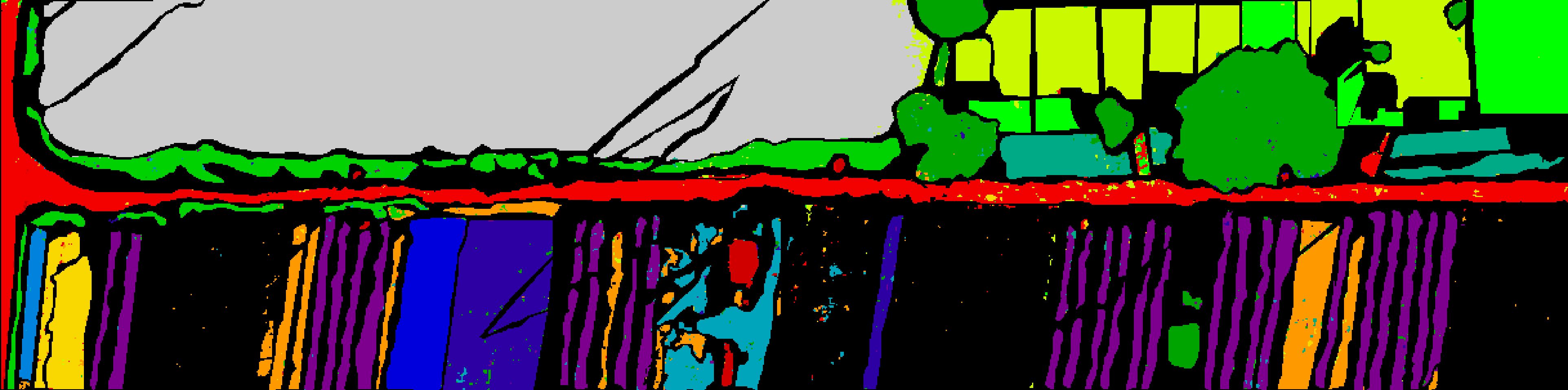}
	\caption*{3D}
    \end{subfigure} 
    \begin{subfigure}{0.11\textwidth}
	\includegraphics[width=0.99\textwidth]{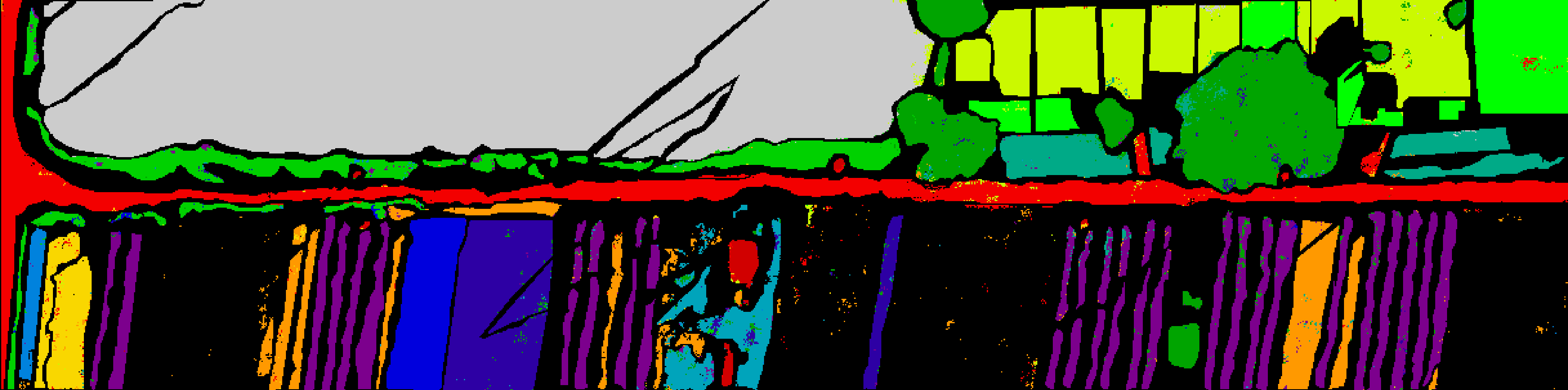}
	\caption*{HSST}
    \end{subfigure}
    \begin{subfigure}{0.11\textwidth}
	\includegraphics[width=0.99\textwidth]{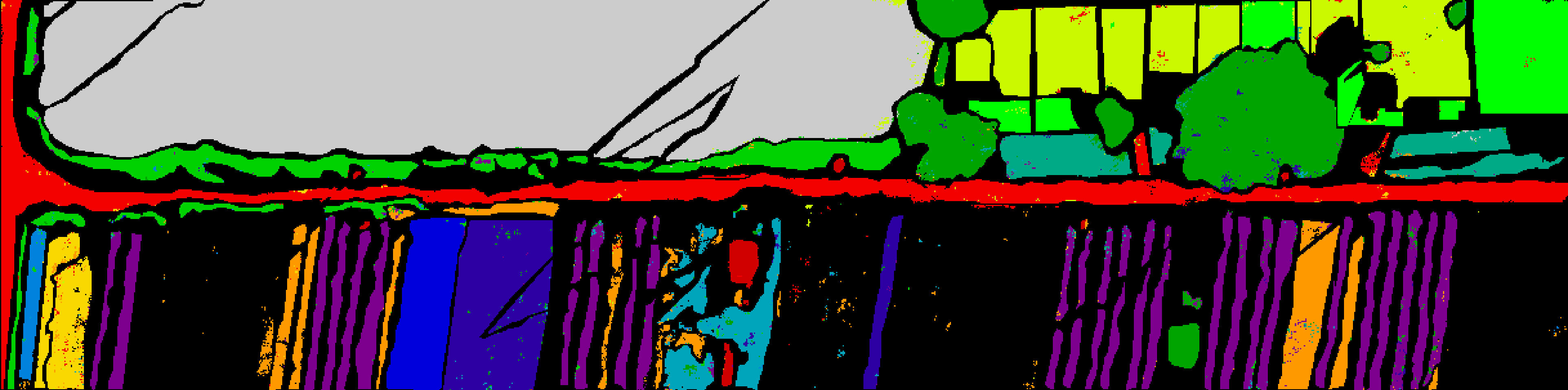}
	\caption*{SST}
    \end{subfigure}
    \begin{subfigure}{0.11\textwidth}
	\includegraphics[width=0.99\textwidth]{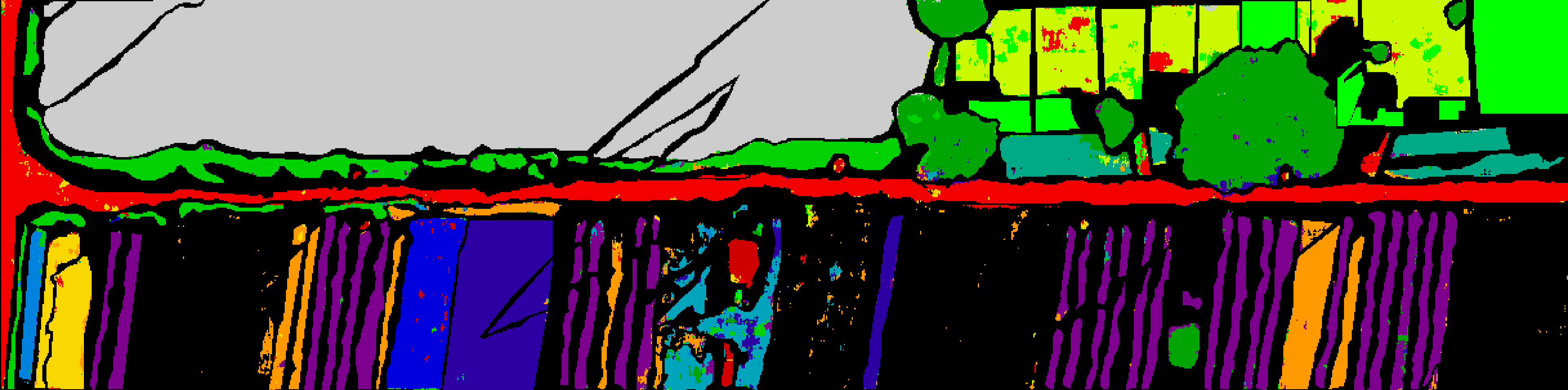}
	\caption*{PF}
    \end{subfigure}
    \begin{subfigure}{0.11\textwidth}
	\includegraphics[width=0.99\textwidth]{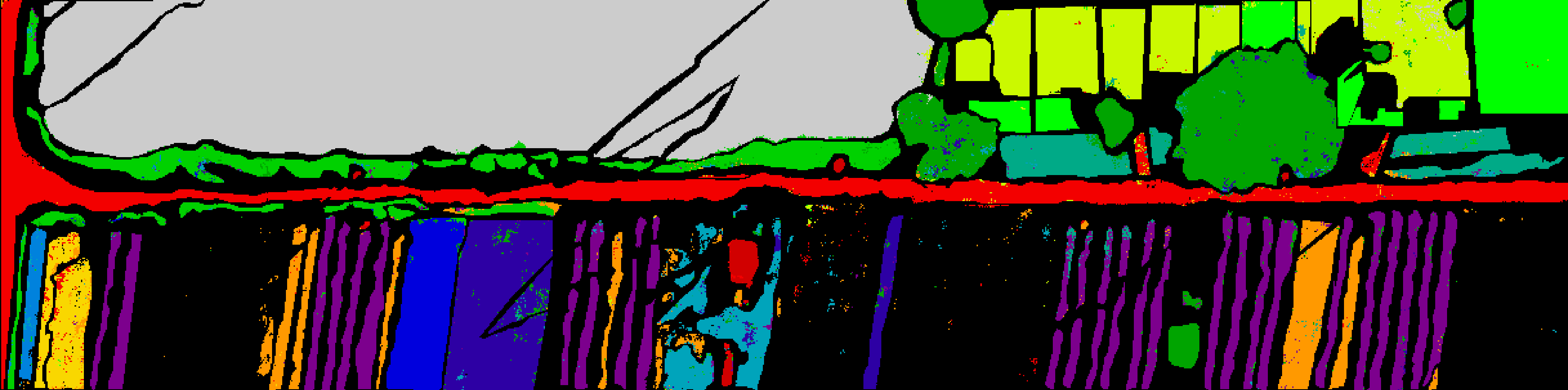}
	\caption*{WF}
    \end{subfigure}
    \begin{subfigure}{0.11\textwidth}
	\includegraphics[width=0.99\textwidth]{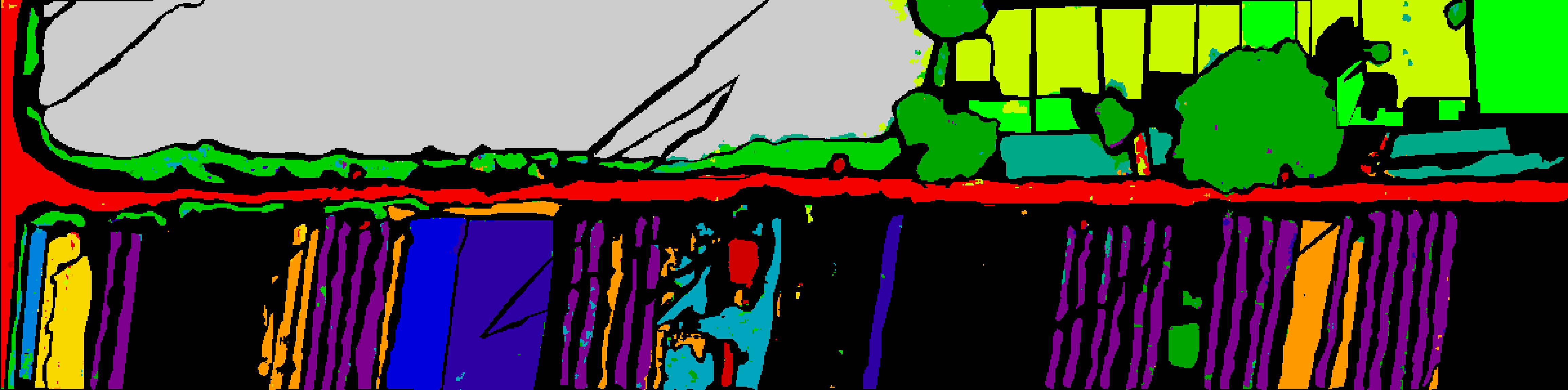}
	\caption*{MSST}
    \end{subfigure}
    \caption{\textbf{HC dataset:} Classification maps comparing baseline models with MSST.}
    \label{HCF}
    \vspace{-.5cm}
\end{figure}

\textbf{HC} Results for HC are presented in Table \ref{HCT} and Fig. \ref{HCF}. \textbf{Observation:} MSST achieves the highest OA (99.27\%) and AA (98.67\%), while requiring fewer parameters than many baselines. The classification maps confirm reduced noise and clearer class separations.

\begin{table}[!hbt]
    \centering
    \caption{\textbf{HC dataset:} Per-class classification results with OA, AA, and $\kappa$.}
        \resizebox{\columnwidth}{!}{\begin{tabular}{c|cccccccc} \hline 
        \textbf{Class} & \textbf{2DCNN} & \textbf{HCNN} & \textbf{3DCNN} & \textbf{HSST} & \textbf{SST} & \textbf{PF} & \textbf{WF} & \textbf{MSST} \\ \hline 
            1 & 98.3234 & 98.7169 & 98.8286 & 98.5693 & 99.1416 & 98.4218 & 98.7705 & \textbf{99.4501} \\
            2 & 98.7167 & 96.8620 & 97.9607 & 96.8445 & 97.4334 & 98.0311 & 96.8884 & \textbf{99.7011} \\
            3 & 95.8584 & 98.3667 & 97.4722 & 97.8417 & 97.9972 & 99.4166 & 95.1390 & \textbf{99.9611} \\
            4 & 99.5143 & 99.4023 & 99.0661 & \textbf{99.8879} & 99.5143 & 95.6667 & 98.2069 & 99.2528 \\
            5 & 92.5000 & 96.1666 & 98.3333 & 97.0000 & 97.0000 & 94.0000 & 89.1666 & \textbf{99.1666} \\
            6 & 80.1412 & 89.7175 & \textbf{95.4986} & 80.2736 & 87.5551 & 67.5639 & 84.4218 & 93.2480 \\
            7 & 93.6991 & 98.2723 & 97.3238 & 97.0867 & 94.6476 & 90.3794 & 92.8184 & \textbf{98.8143} \\
            8 & 95.0828 & 86.7393 & \textbf{98.1087} & 95.2386 & 96.0841 & 94.1039 & 95.4611 & 98.0976 \\
            9 & 97.3806 & 96.6201 & 97.5496 & 94.2120 & 97.0849 & 96.2822 & 93.3248 & \textbf{98.4579} \\
            10 & 98.4024 & 99.5625 & 99.2963 & 98.2312 & 98.3453 & \textbf{99.9239} & 98.7828 & 99.6386 \\
            11 & 98.6162 & 98.9710 & \textbf{99.5742} & 97.8829 & 97.9657 & 86.6351 & 95.9314 & \textbf{99.5505} \\
            12 & 96.4130 & 96.7391 & 97.1195 & 94.6195 & 87.5000 & 93.9673 & 82.1739 & \textbf{99.4021} \\
            13 & 88.7231 & 86.8802 & 92.4967 & 92.0798 & 89.5787 & 88.2404 & 84.4888 & \textbf{97.6305} \\
            14 & 98.1034 & 97.6293 & 95.7435 & 97.5646 & 97.1228 & 96.6810 & 97.5754 & \textbf{99.1918} \\
            15 & 90.8450 & 88.0281 & 89.7887 & 83.6267 & 92.9577 & 86.6197 & 85.2112 & \textbf{97.3591} \\ 
            16 & 99.8938 & 99.8222 & 99.4190 & 99.7374 & 99.5702 & \textbf{99.9071} & 99.6206 & \textbf{99.8328} \\ \hline 
        \textbf{Para} & 1078336 & 1380480 & 15808224 & 836816 & 836816 & 16680662 & 2667408 & 2114128 \\ \hline
        \textbf{Train (s)} & 69.87 & 86.94 & 119.54 & 308.96 & 309.88 & 3185.62 & 305.70 & 232.86 \\ \hline 
        \textbf{Test (s)} & 9.93 & 10.47 & 10.35 & 58.06 & 53.79 & 384.09 & 57.25 & 52.85 \\ \hline 
        \textbf{Flops} & 2670080 & 18778112 & 15732736 & 110592 & 110592 & 15730688 & 167936 & 1999872 \\ \hline 
        \textbf{$\kappa$} & 97.2334 & 96.7964 & 97.9417 & 97.0946 & 97.2897 & 95.7417 & 96.2026 & \textbf{99.1413} \\ \hline
        \textbf{OA} & 97.6367 & 97.2616 & 98.2402 & 97.5179 & 97.6849 & 96.3623 & 96.7568 & \textbf{99.2661} \\ \hline
        \textbf{AA} & 95.1383 & 95.5310 & 97.0987 & 95.0435 & 95.5937 & 92.8650 & 92.9988 & \textbf{98.6722} \\ \hline
    \end{tabular}}
    \label{HCT}
    \vspace{-.2cm}
\end{table}

\begin{table}[!hbt]
    \centering
    \caption{\textbf{HH dataset:} Per-class classification results with OA, AA, and $\kappa$.}
        \resizebox{\columnwidth}{!}{\begin{tabular}{c|cccccccc} \hline 
        \textbf{Class} & \textbf{2DCNN} & \textbf{HCNN} & \textbf{3DCNN} & \textbf{HSST} & \textbf{SST} & \textbf{PF} & \textbf{WF} & \textbf{MSST} \\ \hline 
            1 & 98.8461 & 99.4871 & 99.2877 & 98.6324 & 99.7008 & 97.9202 & 99.2877 & \textbf{99.8005} \\
            2 & 91.6856 & 92.5968 & 93.9066 & 89.8633 & 88.3826 & 72.3234 & 87.6423 & \textbf{98.0068} \\
            3 & 94.3726 & 96.6730 & 98.0661 & \textbf{98.2036} & 97.0030 & 97.6079 & 91.9347 & 97.9561 \\
            4 & 99.4941 & 99.6153 & 99.7623 & 99.7403 & 99.4806 & 99.7329 & 99.6852 & \textbf{99.8211} \\
            5 & 87.1341 & 86.5873 & 79.0286 & 85.2685 & 91.5406 & 82.1807 & 90.5114 & \textbf{94.1138} \\
            6 & 98.2943 & 98.7656 & 98.7252 & 97.5492 & 98.4604 & 97.9487 & 99.2279 & \textbf{99.2279} \\
            7 & 86.7988 & 94.0673 & 90.5741 & 95.0215 & \textbf{97.0046} & 88.6574 & 95.1128 & 96.8138 \\
            8 & 72.1756 & 65.6635 & 73.6063 & 87.4691 & 80.0690 & 40.4538 & 79.4277 & \textbf{98.9639} \\
            9 & 98.7982 & 99.1865 & 98.6319 & 98.5949 & 99.3344 & 97.7999 & 99.1310 & \textbf{99.7781} \\
            10 & 90.3985 & 92.7384 & 87.2357 & 90.5599 & 85.3961 & 90.0758 & 86.4934 & \textbf{97.9506} \\
            11 & 94.0257 & 90.5574 & 85.7998 & 95.0789 & \textbf{95.8234} & 93.4083 & 92.7728 & 93.6081 \\
            12 & 81.5501 & 84.2305 & 89.3902 & 88.9881 & 85.0792 & 89.4125 & 94.7286 & \textbf{95.2200} \\
            13 & 91.4252 & 89.1416 & 88.9994 & 94.8995 & 94.0287 & 92.4293 & 87.7376 & \textbf{98.1606} \\
            14 & 95.0244 & 95.8401 & 91.5443 & 94.5078 & \textbf{97.1179} & 95.7313 & 96.3839 & 93.7466 \\
            15 & 74.6506 & 92.8143 & 84.8303 & 96.0079 & 94.8103 & 88.4231 & 93.4131 & \textbf{98.4031} \\ 
            16 & 97.4662 & 88.9562 & 93.6381 & 94.7948 & 98.1823 & 90.9666 & 99.2564 & \textbf{98.7606} \\
            17 & 97.0764 & 95.8803 & 95.9468 & 98.2724 & 92.6245 & \textbf{99.4019} & 41.4617 & 99.3355 \\
            18 & 95.0870 & 97.3880 & 95.2736 & 97.0771 & 95.8333 & \textbf{99.0671} & 96.3308 & 98.0099 \\
            19 & 94.4903 & 95.5922 & 93.7098 & 96.7860 & 95.0872 & 94.6740 & \textbf{97.3829} & 93.1129 \\
            20 & 88.0091 & 91.1072 & 94.7217 & 94.9512 & 97.9345 & 95.9265 & 95.2954 & \textbf{99.0820} \\
            21 & 81.9277 & 60.0903 & 73.3433 & 76.9578 & 62.1987 & 38.4036 & 93.2228 & \textbf{95.3313} \\ 
            22 & 77.8712 & 88.6633 & 93.4653 & 97.1782 & \textbf{97.5742} & 73.2673 & 95.4455 & 94.5049 \\ \hline 
        \textbf{Para} & 621126 & 922502 & 8928486 & 811478 & 811478 & 9412188 & 2615190 & 817174 \\ \hline

        \textbf{Train (s)} & 107.14 & 126.55 & 145.88 & 476.38 & 664.13 & 1655.00 & 451.15 &  588.35 \\ \hline 
        \textbf{Test (s)} & 13.31 & 14.87 & 13.96 & 80.85 & 121.23 & 166.06 & 81.53 & 84.47 \\ \hline 
        \textbf{Flops} & 1505248 & 10578688 & 8852992 & 112128 & 112128 & 8850176 & 169472 & 1062400 \\ \hline 

        \textbf{$\kappa$} & 94.4348 & 95.1452 & 94.7882 & 96.4281 & 96.2205 & 94.1511 & 95.2651 & \textbf{98.1217} \\ \hline
        \textbf{OA} & 95.5985 & 96.1607 & 95.8835 & 97.1755 & 97.0084 & 95.3772 & 96.2549 & \textbf{98.5140} \\ \hline
        \textbf{AA} & 90.3001 & 90.7110 & 90.8858 & 93.9274 & 92.8485 & 87.0824 & 91.4493 & \textbf{97.2595} \\ \hline
    \end{tabular}}
    \label{HHT}
\end{table}
\begin{figure}[!hbt]
    \centering
    \begin{subfigure}{0.11\textwidth}
	\includegraphics[width=0.99\textwidth]{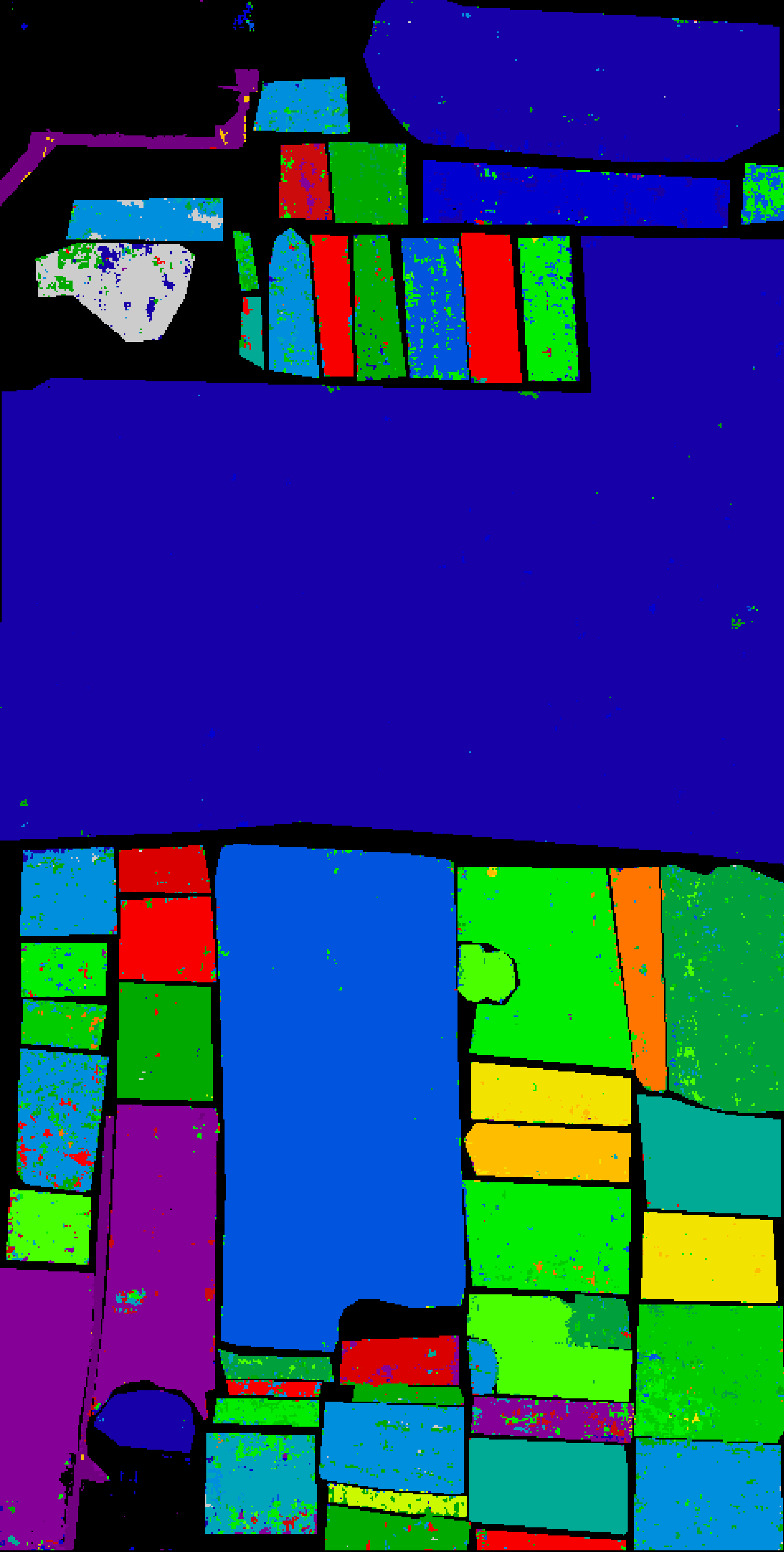}
	\caption*{2D} 
    \end{subfigure}
    \begin{subfigure}{0.11\textwidth}
	\includegraphics[width=0.99\textwidth]{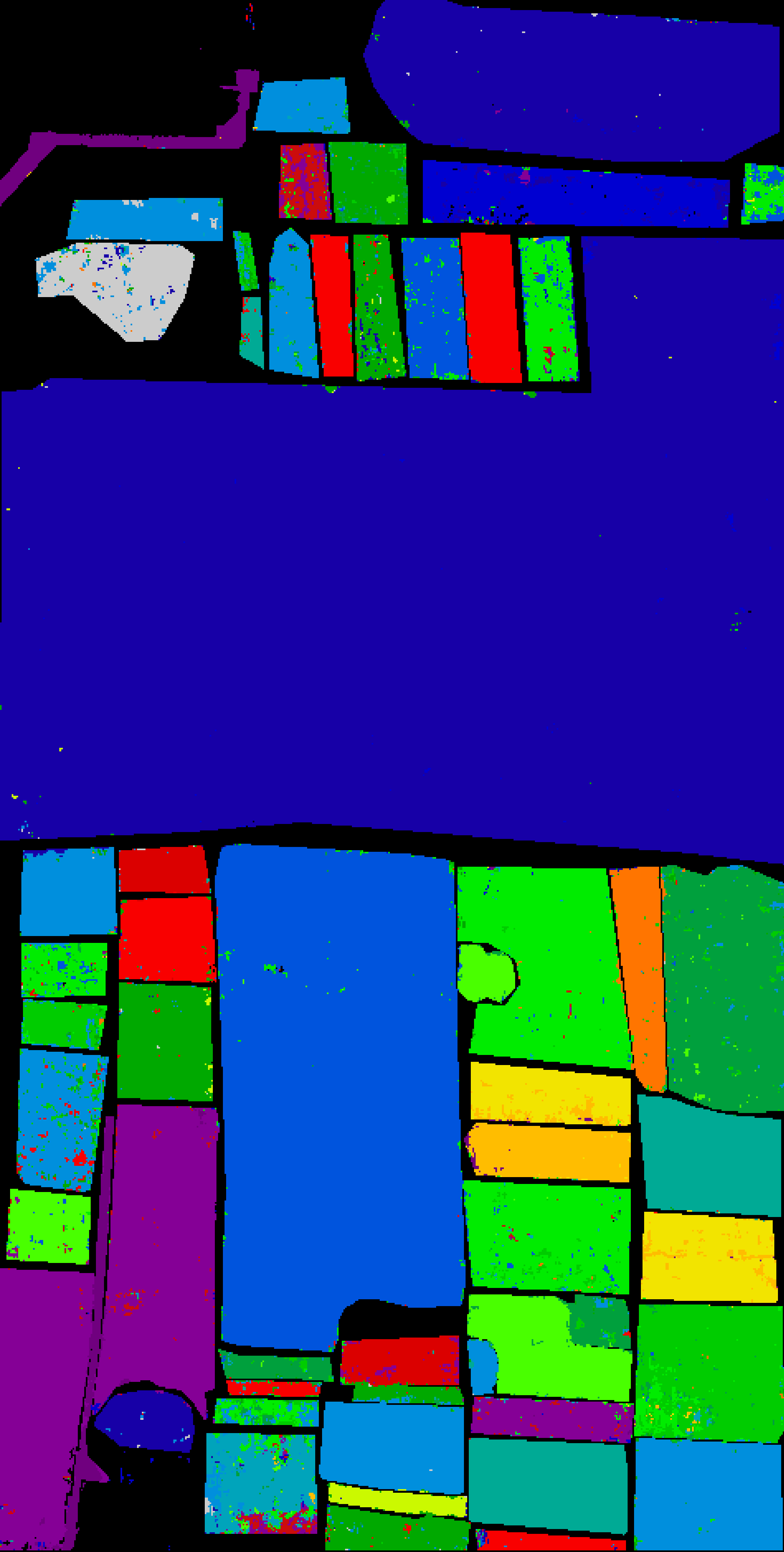}
	\caption*{Hyb}
    \end{subfigure}
    \begin{subfigure}{0.11\textwidth}
	\includegraphics[width=0.99\textwidth]{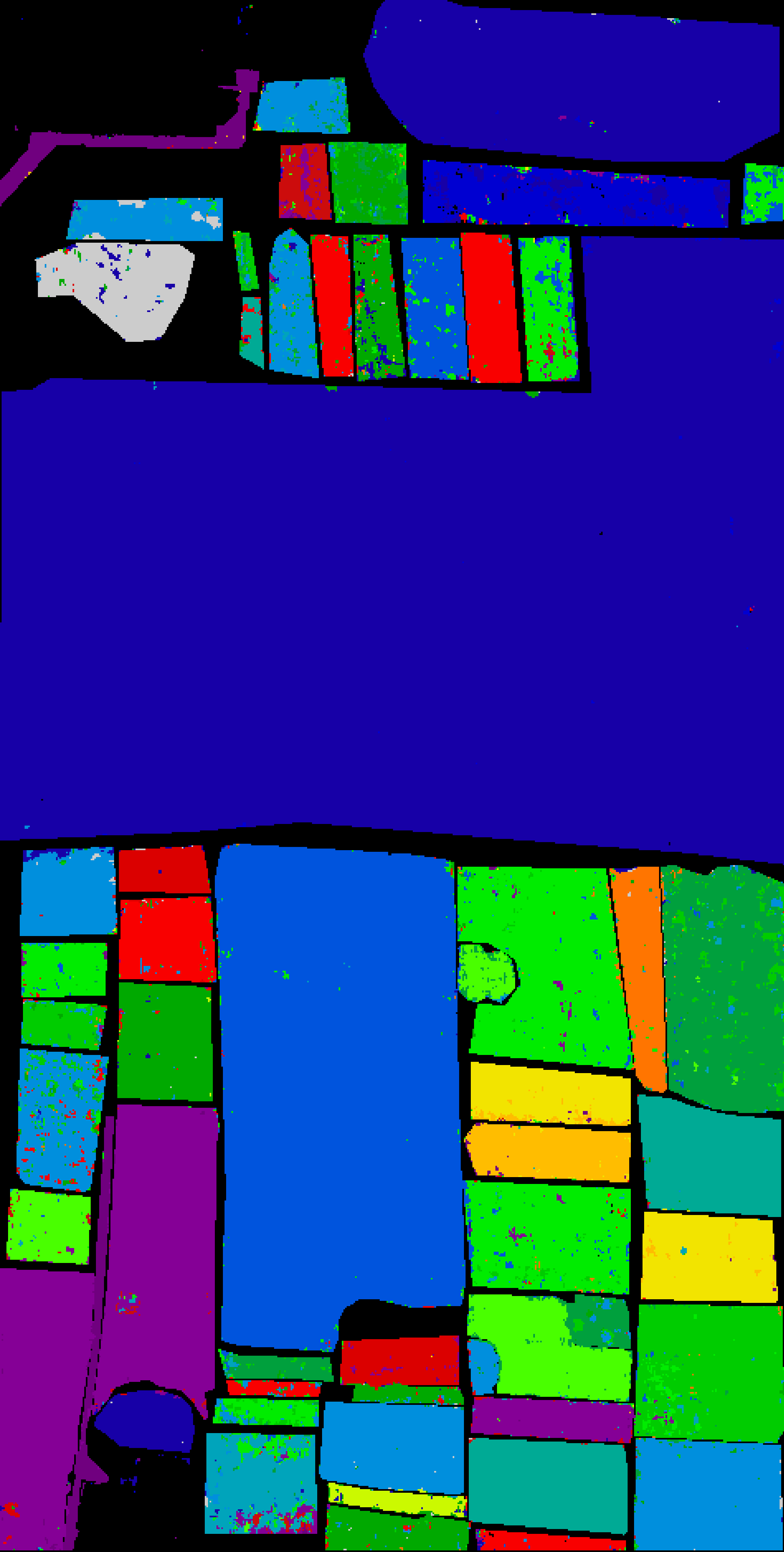}
	\caption*{3D}
    \end{subfigure} 
    \begin{subfigure}{0.11\textwidth}
	\includegraphics[width=0.99\textwidth]{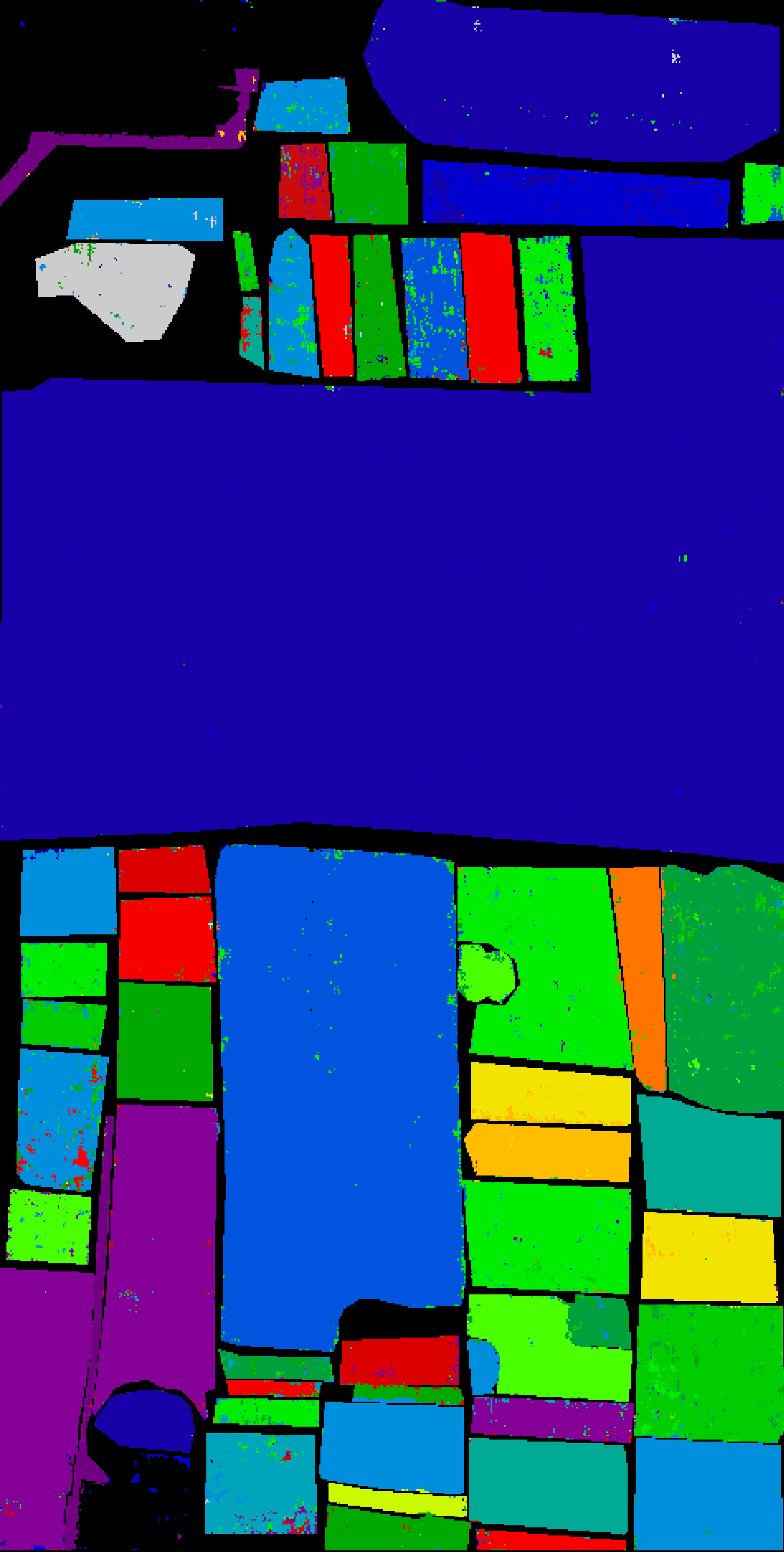}
	\caption*{HSST}
    \end{subfigure}
    \begin{subfigure}{0.11\textwidth}
	\includegraphics[width=0.99\textwidth]{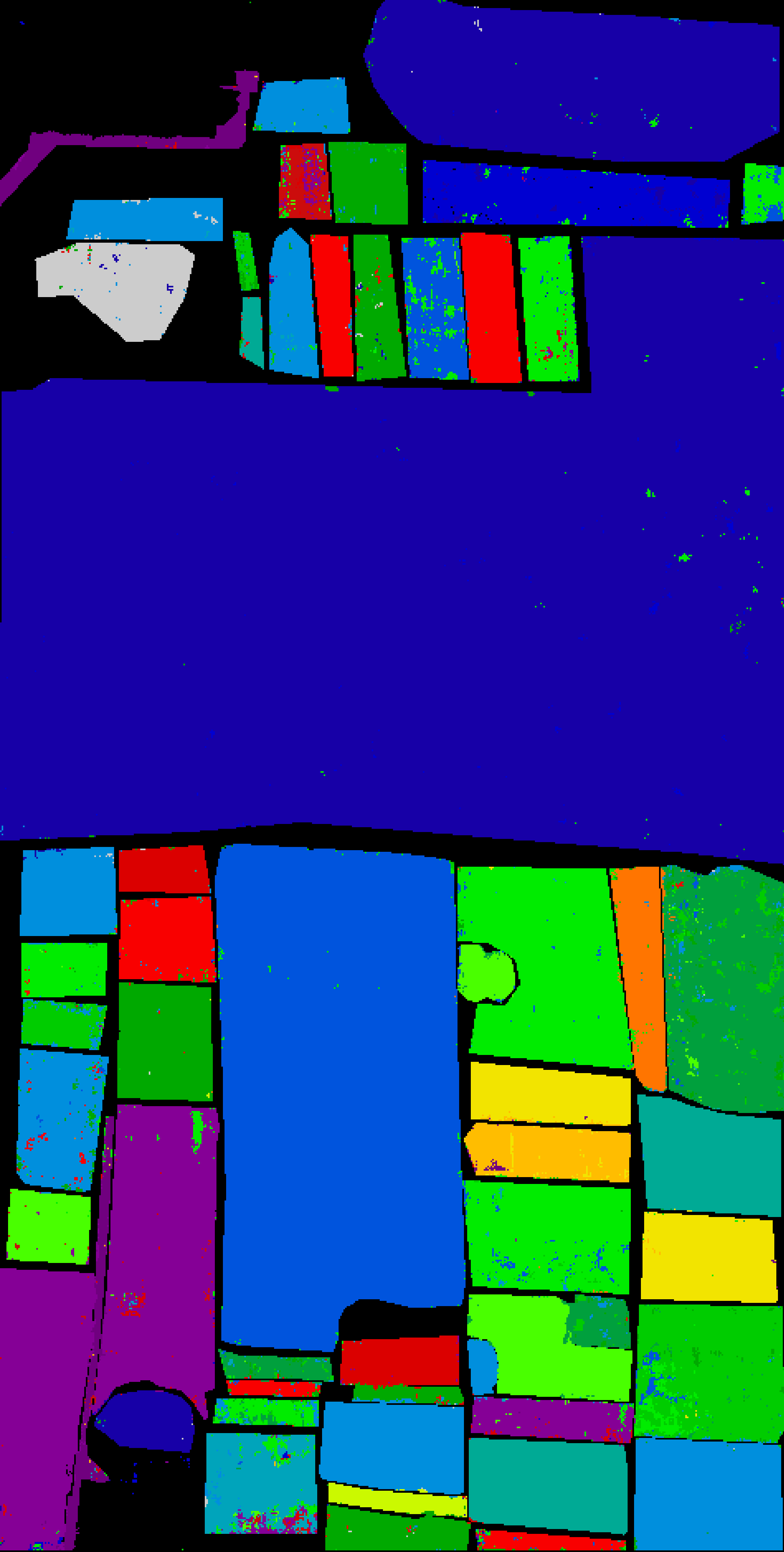}
	\caption*{SST}
    \end{subfigure}
    \begin{subfigure}{0.11\textwidth}
	\includegraphics[width=0.99\textwidth]{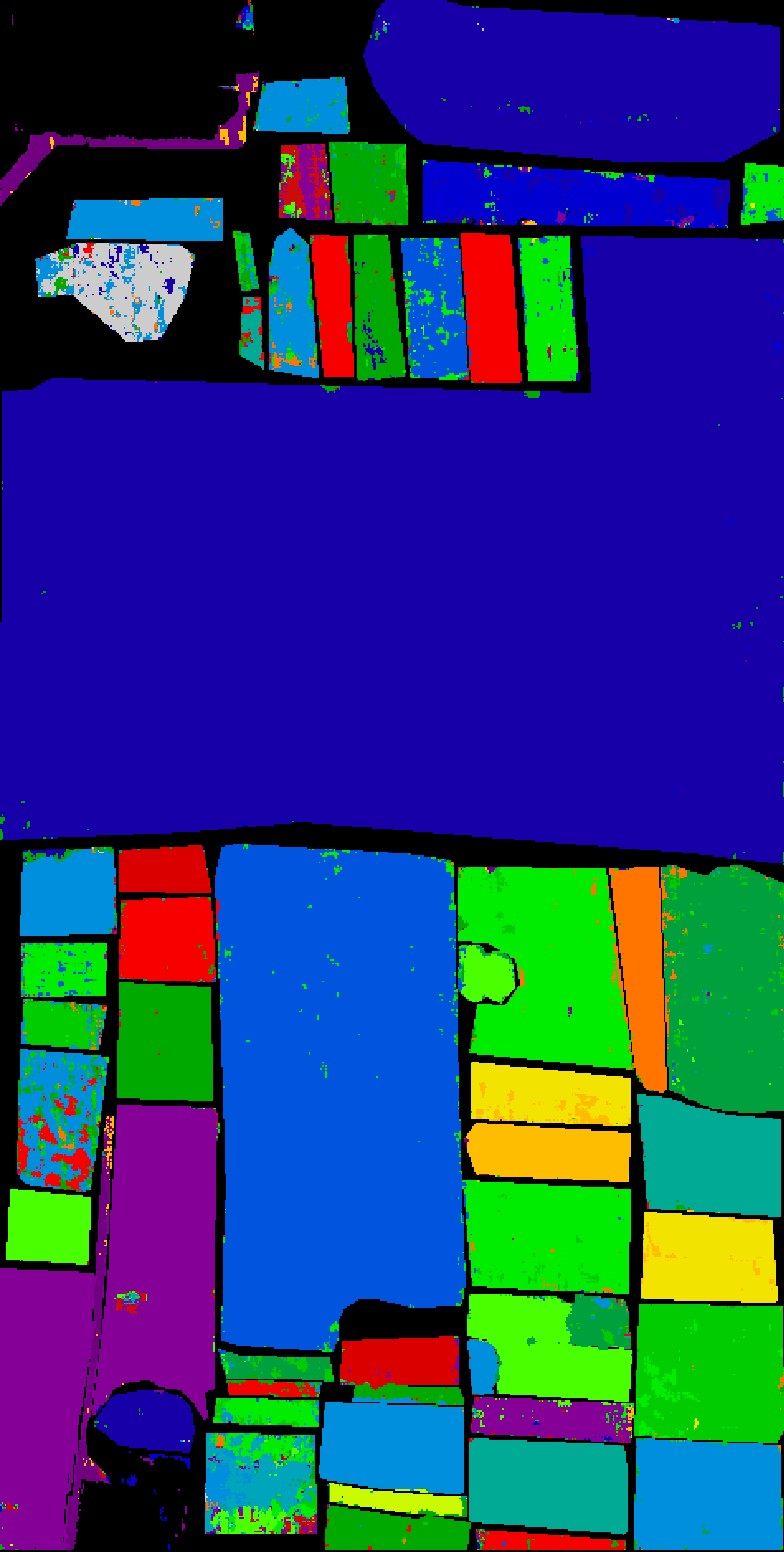}
	\caption*{PF}
    \end{subfigure}
    \begin{subfigure}{0.11\textwidth}
	\includegraphics[width=0.99\textwidth]{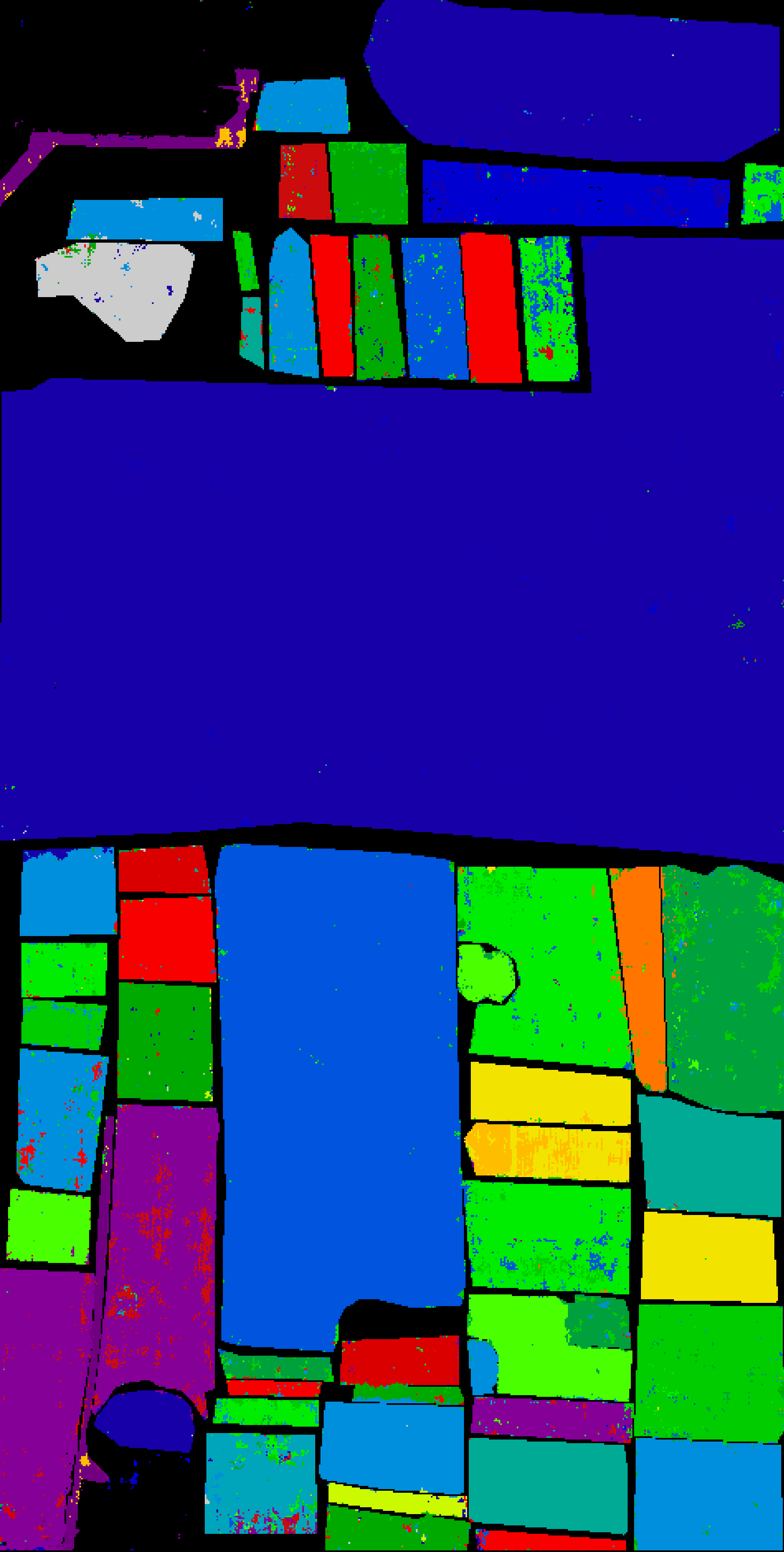}
	\caption*{WF}
    \end{subfigure}
    \begin{subfigure}{0.11\textwidth}
	\includegraphics[width=0.99\textwidth]{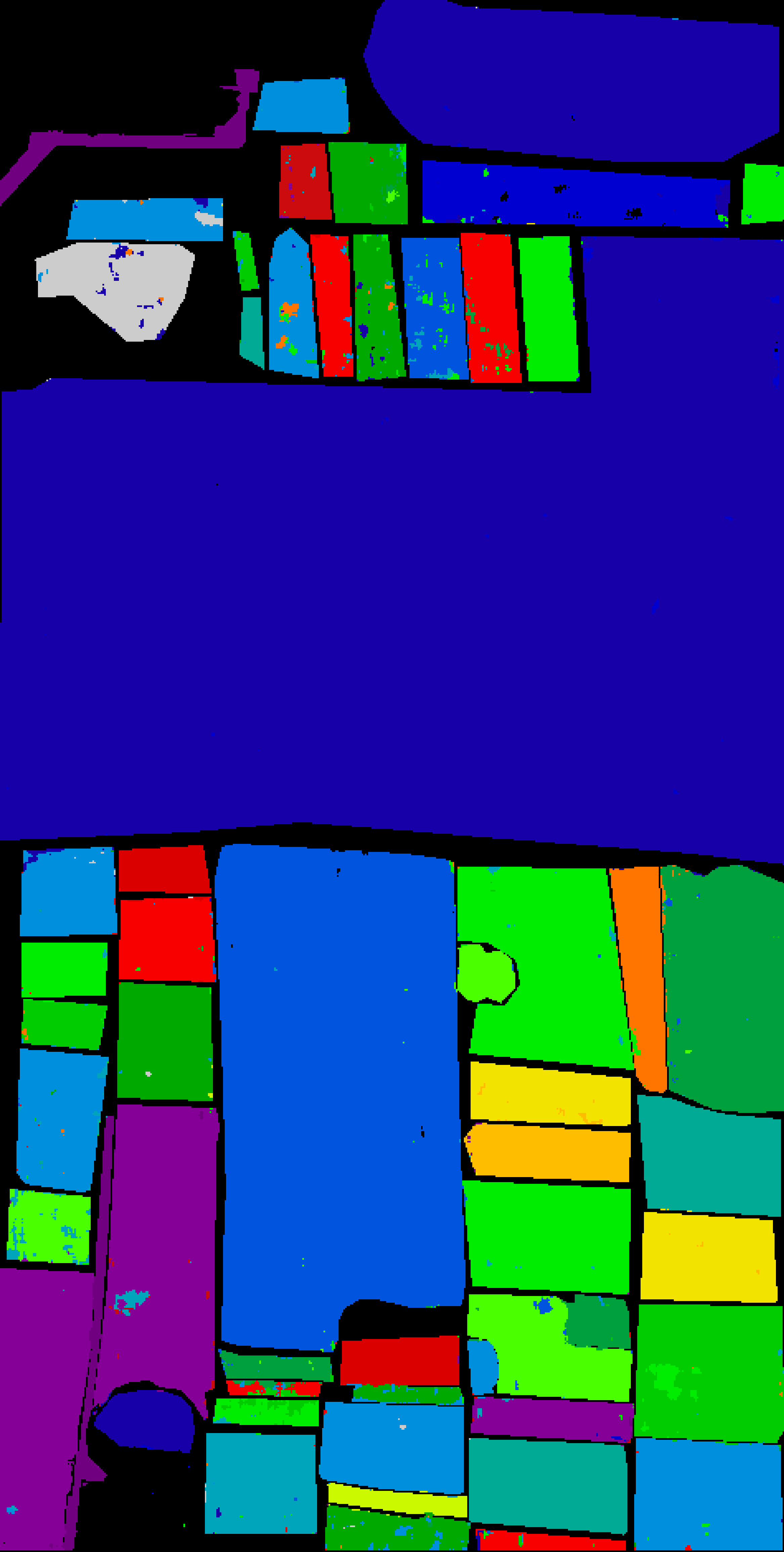}
	\caption*{MSST}
    \end{subfigure}
    \caption{\textbf{HH dataset:} Classification maps, highlighting spatial variability and class-specific performance.}
    \label{HHF}
    \vspace{-.2cm}
\end{figure}

\textbf{HH} Table \ref{HHT} and Fig. \ref{HHF} report results for HH. \textbf{Observation:} MSST achieves the best results across all metrics, particularly in difficult classes (e.g., Class 2 and Class 8). Despite higher accuracy, MSST remains lightweight with only 0.82M parameters.

Across all datasets, MSST consistently outperforms both CNN-based and transformer-based competitors. Importantly, it achieves this with substantially fewer parameters than heavy architectures like PF and 3D CNN. The classification maps further confirm MSST’s ability to capture fine-grained spatial–spectral structures, yielding reduced misclassification and smoother boundaries. This balance of accuracy, efficiency, and structural consistency highlights the practical potential of MSST for real-world HSIC applications.

\section{Conclusions and Future Research Directions}
\label{sec:concl}

This paper proposed MemFormer, a transformer-based architecture for HSIC that integrates a dynamic memory-enhanced attention mechanism with a SSPE. By augmenting standard self-attention with a compact global memory, the proposed framework enables effective modeling of long-range contextual dependencies while mitigating information redundancy. In addition, the SSPE formulation jointly captures spatial continuity and spectral ordering without relying on computationally intensive convolutional encodings.

Experimental evaluations conducted on three widely used hyperspectral benchmark datasets demonstrate that MemFormer achieves superior classification performance compared to representative convolutional, hybrid, and transformer-based approaches. Ablation analyses further validate the individual contributions of the dynamic memory module, the SSPE design, and the memory size configuration, highlighting their roles in balancing representational capacity and efficiency.

Despite these results, several research directions remain open. First, extending the proposed framework to large-scale hyperspectral scenes with complex acquisition conditions poses challenges in memory management and attention efficiency. Second, incorporating domain adaptation and self-supervised learning strategies may improve robustness to sensor variability, atmospheric distortions, and cross-dataset distribution shifts. Third, enhancing the interpretability of memory-driven attention mechanisms represents an important step toward transparent and trustworthy hyperspectral analysis. Finally, future work may explore multimodal extensions of MemFormer by integrating complementary data sources such as LiDAR or SAR to improve discrimination in structurally complex and heterogeneous environments.

\bibliographystyle{IEEEbib}
\bibliography{strings,refs}

@ARTICLE{11392780,
  author={Ahmad, Faiq and Usama, Muhammad and Ghous, Usman and Shehzad, Danish and Mazzara, Manuel and Ahmad, Muhammad},
  journal={IEEE Geoscience and Remote Sensing Letters}, 
  title={A Spiking and Memory-Enhanced State-Space Model for Hyperspectral Image Classification}, 
  year={2026},
  volume={23},
  number={},
  pages={1-5},
  doi={10.1109/LGRS.2026.3663905}}

@ARTICLE{11119702,
  author={Sohail, Saad and Usama, Muhammad and Ghous, Usman and Mazzara, Manuel and Distefano, Salvatore and Ahmad, Muhammad},
  journal={IEEE Geoscience and Remote Sensing Letters}, 
  title={EnergyFormer: Energy Attention With Fourier Embedding for Hyperspectral Image Classification}, 
  year={2025},
  volume={22},
  number={},
  pages={1-5},
  doi={10.1109/LGRS.2025.3596629}}

@ARTICLE{11105087,
  author={Ahmad, Muhammad and Mauro, Francesco and Raza, Rana Aamir and Mazzara, Manuel and Distefano, Salvatore and Khan, Adil Mehmood and Ullo, Silvia Liberata},
  journal={IEEE Journal of Selected Topics in Applied Earth Observations and Remote Sensing}, 
  title={Transformer-Driven Active Transfer Learning for Cross-Hyperspectral Image Classification}, 
  year={2025},
  volume={18},
  number={},
  pages={19635-19648},
  doi={10.1109/JSTARS.2025.3594108}}

@article{Ahmad18072025,
author = {Muhammad Ahmad and Manuel Mazzara and Salvatore Distefano and Adil Mehmood Khan and Xin Wu},
title = {Self-supervised spatial-spectral transformer with Extreme Learning Machine for Hyperspectral Image Classification},
journal = {International Journal of Remote Sensing},
volume = {46},
number = {14},
pages = {5384--5407},
year = {2025},
publisher = {Taylor \& Francis},
doi = {10.1080/01431161.2025.2520049},
URL = {https://doi.org/10.1080/01431161.2025.2520049},
eprint = {https://doi.org/10.1080/01431161.2025.2520049}}

@ARTICLE{11037730,
  author={Sohail, Saad and Usama, Muhammad and Ghous, Usman and Mazzara, Manuel and Ahmad, Muhammad},
  journal={IEEE Geoscience and Remote Sensing Letters}, 
  title={Differential Attention With Enhanced Squeeze-and-Excitation for Hyperspectral Image Classification}, 
  year={2025},
  volume={22},
  number={},
  pages={1-5},
  doi={10.1109/LGRS.2025.3580630}}

@ARTICLE{9767615,
  author={Ahmad, Muhammad and Khan, Adil Mehmood and Mazzara, Manuel and Distefano, Salvatore and Roy, Swalpa Kumar and Wu, Xin},
  journal={IEEE Journal of Selected Topics in Applied Earth Observations and Remote Sensing}, 
  title={Hybrid Dense Network With Attention Mechanism for Hyperspectral Image Classification}, 
  year={2022},
  volume={15},
  number={},
  pages={3948-3957},
  doi={10.1109/JSTARS.2022.3171586}}

@ARTICLE{11222092,
  author={Ahmad, Muhammad and Mazzara, Manuel and Distefano, Salvatore and Khan, Adil Mehmood},
  journal={IEEE Transactions on Geoscience and Remote Sensing}, 
  title={Byte Latent Mamba With State Space and Knowledge Distillation for Hyperspectral Image Classification}, 
  year={2025},
  volume={63},
  number={},
  pages={1-15},
  doi={10.1109/TGRS.2025.3626861}}

@ARTICLE{11226902,
  author={Ahmad, Muhammad and Mazzara, Manuel and Distefano, Salvatore and Khan, Adil Mehmood and Butt, Muhammad Hassaan Farooq and Usama, Muhammad and Hong, Danfeng},
  journal={IEEE Transactions on Emerging Topics in Computing}, 
  title={GraphMamba: Graph Tokenization Mamba for Hyperspectral Image Classification}, 
  year={2025},
  volume={13},
  number={4},
  pages={1510-1521},
  doi={10.1109/TETC.2025.3626943}}

@ARTICLE{11090003,
  author={Ahmad, Muhammad and Mazzara, Manuel and Distefano, Salvatore and Mehmood Khan, Adil and Hassaan Farooq Butt, Muhammad and Hong, Danfeng},
  journal={IEEE Transactions on Neural Networks and Learning Systems}, 
  title={PolicyMamba: Localized Policy Attention With State Space Model for Land Cover Classification}, 
  year={2025},
  volume={36},
  number={10},
  pages={17814-17825},
  doi={10.1109/TNNLS.2025.3586836}}

@article{Ahmad03042025,
author = {Muhammad Ahmad and Muhammad Hassaan Farooq Butt and Muhammad Usama and Hamad Ahmed Altuwaijri and Manuel Mazzara and Salvatore Distefano and Adil Mehmood Khan and},
title = {Multi-head spatial-spectral mamba for hyperspectral image classification},
journal = {Remote Sensing Letters},
volume = {16},
number = {4},
pages = {339--353},
year = {2025},
publisher = {Taylor \& Francis},
doi = {10.1080/2150704X.2025.2461330},
URL = {https://doi.org/10.1080/2150704X.2025.2461330},
eprint = {https://doi.org/10.1080/2150704X.2025.2461330}}

@Article{rs13122275,
AUTHOR = {Ahmad, Muhammad and Mazzara, Manuel and Distefano, Salvatore},
TITLE = {Regularized CNN Feature Hierarchy for Hyperspectral Image Classification},
JOURNAL = {Remote Sensing},
VOLUME = {13},
YEAR = {2021},
NUMBER = {12},
ARTICLE-NUMBER = {2275},
URL = {https://www.mdpi.com/2072-4292/13/12/2275},
ISSN = {2072-4292},
DOI = {10.3390/rs13122275}}

@ARTICLE{9645266,
  author={Ahmad, Muhammad and Shabbir, Sidrah and Roy, Swalpa Kumar and Hong, Danfeng and Wu, Xin and Yao, Jing and Khan, Adil Mehmood and Mazzara, Manuel and Distefano, Salvatore and Chanussot, Jocelyn},
  journal={IEEE Journal of Selected Topics in Applied Earth Observations and Remote Sensing}, 
  title={Hyperspectral Image Classification—Traditional to Deep Models: A Survey for Future Prospects}, 
  year={2022},
  volume={15},
  number={},
  pages={968-999},
  doi={10.1109/JSTARS.2021.3133021}}

@ARTICLE{10767233,
  author={Ahmad, Muhammad and Usama, Muhammad and Mazzara, Manuel and Distefano, Salvatore},
  journal={IEEE Geoscience and Remote Sensing Letters}, 
  title={WaveMamba: Spatial-Spectral Wavelet Mamba for Hyperspectral Image Classification}, 
  year={2025},
  volume={22},
  number={},
  pages={1-5},
  doi={10.1109/LGRS.2024.3506034}}

@ARTICLE{10399798,
  author={Ahmad, Muhammad and Ghous, Usman and Usama, Muhammad and Mazzara, Manuel},
  journal={IEEE Geoscience and Remote Sensing Letters}, 
  title={WaveFormer: Spectral–Spatial Wavelet Transformer for Hyperspectral Image Classification}, 
  year={2024},
  volume={21},
  number={},
  pages={1-5},
  doi={10.1109/LGRS.2024.3353909}}

@ARTICLE{9307220,
  author={Ahmad, Muhammad and Khan, Adil Mehmood and Mazzara, Manuel and Distefano, Salvatore and Ali, Mohsin and Sarfraz, Muhammad Shahzad},
  journal={IEEE Geoscience and Remote Sensing Letters}, 
  title={A Fast and Compact 3-D CNN for Hyperspectral Image Classification}, 
  year={2022},
  volume={19},
  number={},
  pages={1-5},
  doi={10.1109/LGRS.2020.3043710}}

@article{AHMAD2025130428,
title = {A comprehensive survey for Hyperspectral Image Classification: The evolution from conventional to transformers and Mamba models},
journal = {Neurocomputing},
volume = {644},
pages = {130428},
year = {2025},
issn = {0925-2312},
doi = {https://doi.org/10.1016/j.neucom.2025.130428},
url = {https://www.sciencedirect.com/science/article/pii/S0925231225011002},
author = {Muhammad Ahmad and Salvatore Distefano and Adil Mehmood Khan and Manuel Mazzara and Chenyu Li and Hao Li and Jagannath Aryal and Yao Ding and Gemine Vivone and Danfeng Hong}}

@ARTICLE{9903062,
  author={Ahmad, Muhammad and Ghous, Usman and Hong, Danfeng and Khan, Adil Mehmood and Yao, Jing and Wang, Shaohua and Chanussot, Jocelyn},
  journal={IEEE Transactions on Geoscience and Remote Sensing}, 
  title={A Disjoint Samples-Based 3D-CNN With Active Transfer Learning for Hyperspectral Image Classification}, 
  year={2022},
  volume={60},
  number={},
  pages={1-16},
  doi={10.1109/TGRS.2022.3209182}}

@ARTICLE{10604879,
  author={Ahmad, Muhammad and Usama, Muhammad and Khan, Adil Mehmood and Distefano, Salvatore and Altuwaijri, Hamad Ahmed and Mazzara, Manuel},
  journal={IEEE Geoscience and Remote Sensing Letters}, 
  title={Spatial–Spectral Transformer With Conditional Position Encoding for Hyperspectral Image Classification}, 
  year={2024},
  volume={21},
  number={},
  pages={1-5},
  doi={10.1109/LGRS.2024.3431188}}

@ARTICLE{10685113,
  author={Ahmad, Muhammad and Usama, Muhammad and Mazzara, Manuel and Distefano, Salvatore and Altuwaijri, Hamad Ahmed and Ullo, Silvia Liberata},
  journal={IEEE Journal of Selected Topics in Applied Earth Observations and Remote Sensing}, 
  title={Fusing Transformers in a Tuning Fork Structure for Hyperspectral Image Classification Across Disjoint Samples}, 
  year={2024},
  volume={17},
  number={},
  pages={18167-18181},
  doi={10.1109/JSTARS.2024.3465831}}

@article{ahmad2024multihead,
author = {Ahmad, Muhammad and Butt, Muhammad Hassaan Farooq and Usama, Muhammad and Altuwaijri, Hamad Ahmed and Mazzara, Manuel and Distefano, Salvatore and Adil Mehmood Khan},
title = {Multi-head spatial-spectral mamba for hyperspectral image classification},
journal = {Remote Sensing Letters},
volume = {16},
number = {4},
pages = {15--29},
year = {2025},
publisher = {Taylor \& Francis},
doi = {10.1080/2150704X.2025.2461330},
URL = {https://doi.org/10.1080/2150704X.2025.2461330},
eprint = {https://doi.org/10.1080/2150704X.2025.2461330}}

@ARTICLE{10955699,
  author={Ahmad, Muhammad and Mazzara, Manuel and Distefano, Salvatore and Khan, Adil Mehmood and Ullo, Silvia Liberata},
  journal={IEEE Journal of Selected Topics in Applied Earth Observations and Remote Sensing}, 
  title={DiffFormer: a Differential Spatial-Spectral Transformer for Hyperspectral Image Classification}, 
  year={2025},
  volume={},
  number={},
  pages={1-11},
  doi={10.1109/JSTARS.2025.3558889}}

@article{li2024casformer,
  title={CasFormer: Cascaded transformers for fusion-aware computational hyperspectral imaging},
  author={Li, Chenyu and Zhang, Bing and Hong, Danfeng and Zhou, Jun and Vivone, Gemine and Li, Shutao and Chanussot, Jocelyn},
  journal={Information Fusion},
  volume={108},
  pages={102408},
  year={2024},
  publisher={Elsevier}
}

@article{yu2024hypersinet,
  title={Hypersinet: A synergetic interaction network combined with convolution and transformer for hyperspectral image classification},
  author={Yu, Qixing and Wei, Weibo and Li, Dantong and Pan, Zhenkuan and Li, Chenyu and Hong, Danfeng},
  journal={IEEE Transactions on Geoscience and Remote Sensing},
  year={2024},
  publisher={IEEE}
}

@ARTICLE{10506482,
  author={Sun, Le and Zhang, Hang and Zheng, Yuhui and Wu, Zebin and Ye, Zhonglin and Zhao, Haixing},
  journal={IEEE Transactions on Geoscience and Remote Sensing}, 
  title={MASSFormer: Memory-Augmented Spectral-Spatial Transformer for Hyperspectral Image Classification}, 
  year={2024},
  volume={62},
  number={},
  pages={1-15},
  doi={10.1109/TGRS.2024.3392264}}

@ARTICLE{10677405,
  author={Huang, Lingbo and Chen, Yushi and He, Xin},
  journal={IEEE Transactions on Geoscience and Remote Sensing}, 
  title={Foundation Model-Based Spectral–Spatial Transformer for Hyperspectral Image Classification}, 
  year={2024},
  volume={62},
  number={},
  pages={1-25},
  doi={10.1109/TGRS.2024.3456129}}

@ARTICLE{10472541,
  author={Zhao, Zhuoyi and Xu, Xiang and Li, Shutao and Plaza, Antonio},
  journal={IEEE Transactions on Geoscience and Remote Sensing}, 
  title={Hyperspectral Image Classification Using Groupwise Separable Convolutional Vision Transformer Network}, 
  year={2024},
  volume={62},
  number={},
  pages={1-17},
  doi={10.1109/TGRS.2024.3377610}}

@ARTICLE{10681622,
  author={Ahmad, Muhammad and Butt, Muhammad Hassaan Farooq and Mazzara, Manuel and Distefano, Salvatore and Khan, Adil Mehmood and Altuwaijri, Hamad Ahmed},
  journal={IEEE Journal of Selected Topics in Applied Earth Observations and Remote Sensing}, 
  title={Pyramid Hierarchical Spatial-Spectral Transformer for Hyperspectral Image Classification}, 
  year={2024},
  volume={17},
  number={},
  pages={17681-17689},
  doi={10.1109/JSTARS.2024.3461851}}

@ARTICLE{10443948,
  author={Jia, Sen and Wang, Yifan and Jiang, Shuguo and He, Ruyan},
  journal={IEEE Transactions on Geoscience and Remote Sensing}, 
  title={A Center-Masked Transformer for Hyperspectral Image Classification}, 
  year={2024},
  volume={62},
  number={},
  pages={1-16},
  doi={10.1109/TGRS.2024.3369075}}

@ARTICLE{10681548,
  author={Liu, Wei and Prasad, Saurabh and Crawford, Melba},
  journal={IEEE Transactions on Geoscience and Remote Sensing}, 
  title={Investigation of Hierarchical Spectral Vision Transformer Architecture for Classification of Hyperspectral Imagery}, 
  year={2024},
  volume={62},
  number={},
  pages={1-19},
  doi={10.1109/TGRS.2024.3462374}}

@ARTICLE{10599231,
  author={Shi, Cuiping and Yue, Shuheng and Wang, Liguo},
  journal={IEEE Transactions on Geoscience and Remote Sensing}, 
  title={Attention Head Interactive Dual Attention Transformer for Hyperspectral Image Classification}, 
  year={2024},
  volume={62},
  number={},
  pages={1-20},
  doi={10.1109/TGRS.2024.3427769}}

@ARTICLE{10419118,
  author={Yu, Lili and Zhang, Xubing and Wang, Kai},
  journal={IEEE Transactions on Geoscience and Remote Sensing}, 
  title={CMAAC: Combining Multiattention and Asymmetric Convolution Global Learning Framework for Hyperspectral Image Classification}, 
  year={2024},
  volume={62},
  number={},
  pages={1-18},
  doi={10.1109/TGRS.2024.3361555}}

@ARTICLE{10711882,
  author={Kumar Mp, Pavan and Tu, Zhe-Xiang and Chen, Hsu-Chi and Chen, Kun-Chih},
  journal={IEEE Transactions on Instrumentation and Measurement}, 
  title={Mitigating Negative Transfer Learning in Source Free-Unsupervised Domain Adaptation for Rotating Machinery Fault Diagnosis}, 
  year={2024},
  volume={73},
  number={},
  pages={1-16},
  doi={10.1109/TIM.2024.3476610}}

@ARTICLE{10745620,
  author={Wan, Xiaoqing and Chen, Feng and Gao, Weizhe and He, Yupeng and Liu, Hui and Li, Zhize},
  journal={IEEE Journal of Selected Topics in Applied Earth Observations and Remote Sensing}, 
  title={Efficient Spectral-Spatial Fusion With Multiscale and Adaptive Attention for Hyperspectral Image Classification}, 
  year={2025},
  volume={18},
  number={},
  pages={1196-1211},
  doi={10.1109/JSTARS.2024.3492351}}

@article{hu2015deep,
  title={Deep convolutional neural networks for hyperspectral image classification},
  author={Hu, Wei and Huang, Yangyu and Wei, Li and Zhang, Fan and Li, Hengchao},
  journal={Journal of Sensors},
  volume={2015},
  year={2015},
  publisher={Hindawi}
}

@article{hamida2018deep,
  title={3-D Deep Learning Approach for Remote Sensing Image Classification},
  author={Ben Hamida, Amina and Benoit, Alexandre and Lambert, Patrick and Ben Amar, Chokri},
  journal={IEEE Transactions on geoscience and remote sensing},
  volume={56},
  number={8},
  pages={4420--4434},
  year={2018},
  publisher={IEEE}
}

@article{paoletti2018new,
  title={A new deep convolutional neural network for fast hyperspectral image classification},
  author={Paoletti, ME and Haut, JM and Plaza, J and Plaza, A},
  journal={ISPRS journal of photogrammetry and remote sensing},
  volume={145},
  pages={120--147},
  year={2018},
  publisher={Elsevier}
}

@ARTICLE{10884842,
  author={Wu, Jinbin and Zhao, Jiankang and Long, Haihui},
  journal={IEEE Transactions on Geoscience and Remote Sensing}, 
  title={Advanced Hyperspectral Image Classification via Spectral–Spatial Redundancy Reduction and TokenLearner-Enhanced Transformer}, 
  year={2025},
  volume={63},
  number={},
  pages={1-12},
  doi={10.1109/TGRS.2025.3541879}}

@ARTICLE{10820058,
  author={Fang, Yu and Sun, Le and Zheng, Yuhui and Wu, Zebin},
  journal={IEEE Transactions on Image Processing}, 
  title={Deformable Convolution-Enhanced Hierarchical Transformer With Spectral-Spatial Cluster Attention for Hyperspectral Image Classification}, 
  year={2025},
  volume={34},
  number={},
  pages={701-716},
  doi={10.1109/TIP.2024.3522809}}

@article{ZHANG2025111470,
title = {Tensor Transformer for hyperspectral image classification},
journal = {Pattern Recognition},
volume = {163},
pages = {111470},
year = {2025},
issn = {0031-3203},
doi = {https://doi.org/10.1016/j.patcog.2025.111470},
url = {https://www.sciencedirect.com/science/article/pii/S003132032500130X},
author = {Wei-Tao Zhang and Yv Bai and Sheng-Di Zheng and Jian Cui and Zhen-zhen Huang}}

@ARTICLE{10843260,
  author={Xi, Bobo and Zhang, Yun and Li, Jiaojiao and Zheng, Tie and Zhao, Xunfeng and Xu, Haitao and Xue, Changbin and Li, Yunsong and Chanussot, Jocelyn},
  journal={IEEE Transactions on Geoscience and Remote Sensing}, 
  title={MCTGCL: Mixed CNN–Transformer for Mars Hyperspectral Image Classification With Graph Contrastive Learning}, 
  year={2025},
  volume={63},
  number={},
  pages={1-14},
  doi={10.1109/TGRS.2025.3529996}}

@ARTICLE{10035973,
  author={Aburaed, Nour and Alkhatib, Mohammed Q. and Marshall, Stephen and Zabalza, Jaime and Al Ahmad, Hussain},
  journal={IEEE Journal of Selected Topics in Applied Earth Observations and Remote Sensing}, 
  title={A Review of Spatial Enhancement of Hyperspectral Remote Sensing Imaging Techniques}, 
  year={2023},
  volume={16},
  number={},
  pages={2275-2300},
  doi={10.1109/JSTARS.2023.3242048}}

@ARTICLE{11322762,
  author={Li, Dekai and Bhatti, Uzair Aslam and Huang, Mengxing and Bruzzone, Lorenzo and Li, Jiaxin},
  journal={IEEE Transactions on Geoscience and Remote Sensing}, 
  title={HyPyraMamba: A Pyramid Spectral Attention and Mamba-Based Architecture for Robust Hyperspectral Image Classification}, 
  year={2026},
  volume={64},
  number={},
  pages={1-16},
  doi={10.1109/TGRS.2025.3650350}}

@ARTICLE{11321257,
  author={He, Zhijie and Weng, Xiang and Fang, Kai and Li, Yane and Ruan, Yaoping and Feng, Hailin},
  journal={IEEE Journal of Selected Topics in Applied Earth Observations and Remote Sensing}, 
  title={BRSMamba: Boundary-Aware Mamba for Forest and Shrub Segmentation From Diverse Satellite Imagery}, 
  year={2026},
  volume={19},
  number={},
  pages={4607-4628},
  doi={10.1109/JSTARS.2025.3650425}}

@ARTICLE{11361224,
  author={Xie, Huaze and Zhao, Xiaobin and Yin, Junjun and Peng, Jiangtao and Wang, Qingwang and Vivone, Gemine},
  journal={IEEE Transactions on Geoscience and Remote Sensing}, 
  title={Cross-Architecture Contrastive Learning for Few-Shot Hyperspectral Image Classification}, 
  year={2026},
  volume={},
  number={},
  pages={1-1},
  doi={10.1109/TGRS.2026.3656174}}
\end{document}